\documentclass[journal]{IEEEtran}
\usepackage{cite}

\ifCLASSINFOpdf
  \usepackage[pdftex]{graphicx}
\else
\fi
\usepackage[cmex10]{amsmath}
\usepackage{siunitx}

\usepackage{algorithmic}
\usepackage[obeyspaces]{url}

\usepackage{booktabs}

\usepackage{placeins}
\usepackage[hidelinks]{hyperref}
\usepackage{cleveref}
\usepackage[acronym,shortcuts]{glossaries}
\usepackage{xcolor}
\usepackage[ruled, vlined]{algorithm2e}
\usepackage[T1]{fontenc}

\usepackage{circledsteps}

\newcommand{\arxivdisclaimer}{%
\def\thefootnote{}%
\footnote{This work has been submitted to the IEEE for possible publication. Copyright may be transferred without notice, after which this version may no longer be accessible.}%
\addtocounter{footnote}{-1}%
}

\begin{document}

\title{Managing Task Execution for Unknown Workloads in Batteryless IoT: A Hardware-Agnostic Evaluation}


\author{Samer Nasser, Henrique Duarte Moura, Ritesh Kumar Singh, Maarten Weyn, and Jeroen Famaey%
\thanks{This research was funded by the AMBIENT-6G and CORRELATE project. The AMBIENT-6G project received funding from the Smart Networks and Services Joint Undertaking (SNS JU) of the European Union’s Horizon Europe research and innovation programme under Grant Agreement No 101192113. CORRELATE was realized in collaboration with imec, with project support from VLAIO (Flanders Innovation and Entrepreneurship).}%
\thanks{S. Nasser, H. D. Moura, R. K. Singh, M. Weyn, and J. Famaey are with the University of Antwerp - imec, IDLab Research Group, Antwerp, Belgium (e-mail: samer.nasser@uantwerpen.be).}%
}

\newacronym{adr}{ADR}{Adaptive Data Rate}
\newacronym{aimd}{AIMD}{additive increase multiplicative decrease}
\newacronym{ap}{AP}{Approximated Prediction}
\newacronym{ble}{BLE}{Bluetooth Low Energy}
\newacronym{cnn}{CNN}{Convolutional Neural Network}
\newacronym{dl}{DL}{Deep Learning}
\newacronym{dnas}{DNAS}{Differentiable Neural Architecture Search}
\newacronym{dqn}{DQN}{Deep Q-Network}
\newacronym{drl}{DRL}{Deep Reinforcement Learning}
\newacronym{ewma}{EWMA}{exponentially weighted moving average}
\newacronym{har}{HAR}{Human Activity Recognition}
\newacronym{iot}{IoT}{Internet of Things}
\newacronym{iti}{ITI}{inter-task interval}
\newacronym{kcl}{KCL}{Kirchhoff’s current law}
\newacronym{kd}{KD}{Knowledge Distillation}
\newacronym{lstm}{LSTM}{Long Short-Term Memory}
\newacronym{mcu}{MCU}{microcontroller unit}
\newacronym{mdp}{MDP}{Markov Decision Process}
\newacronym{miad}{MIAD}{multiplicative increase additive decrease}
\newacronym{ml}{ML}{Machine Learning}
\newacronym{mqtt}{MQTT}{Message Queuing Telemetry Transport}
\newacronym{nas}{NAS}{Neural Architecture Search}
\newacronym{nfr}{NFR}{Negative Failure Reward}
\newacronym{nir}{NIR}{Negative Inaction Reward}
\newacronym{ota-tinyml}{OTA-TinyML}{Over-the-Air TinyML}
\newacronym{pca}{PCA}{Principal Component Analysis}
\newacronym{pir}{PIR}{Positive Inaction Reward}
\newacronym{pmu}{PMU}{power management unit}
\newacronym{ppo}{PPO}{Proximal Policy Optimization}
\newacronym{ptr}{PTR}{Positive Task Reward}
\newacronym{rl}{RL}{Reinforcement Learning}
\newacronym{se}{SE}{speech enhancement}
\newacronym{st}{ST}{Short-Term}
\newacronym{tinyml}{TinyML}{Tiny Machine Learning}
\newacronym{tinyol}{TinyOL}{TinyML with Online-Learning}
\newacronym{tl}{TL}{Transfer Learning}
\newacronym{zed}{ZED}{Zero-energy devices}
\newacronym{mlp}{MLP}{Multi-Layer Perceptron}
\newacronym{pmic}{PMIC}{Power Management Integrated Circuit}
\newacronym{rtc}{RTC}{Real Time Clock}
\newacronym{req}{$R_\text{eq}$}{equivalent resistance}

\maketitle

\arxivdisclaimer

\begin{abstract}
In recent years, the \ac{iot} paradigm has been shifting toward batteryless, energy-harvesting architectures. Sustaining reliable operation in these systems requires intelligent management of highly volatile stored energy. As edge applications grow in complexity, traditional energy-aware schedulers struggle with unpredictable workloads due to their reliance on static execution thresholds or pre-measured, hardware-specific task profiles. To overcome this, we propose two novel, hardware-agnostic dynamic scheduling strategies treating applications as a "black box," requiring no prior energy information: a model-free \ac{rl} agent and an on-the-fly \ac{ap} method. We evaluate these methods against an adaptive task rate approach (AsTAR) and optimized static thresholds using a custom-built, physically accurate simulation framework driven by real-world solar data and dynamic LoRa transmission profiles. Rather than claiming universal superiority, our analysis exposes the distinct operational trade-offs of each method: the \ac{ap} approach delivers lightweight, near-oracle task throughput; the \ac{rl} agent provides tunable survival-execution balancing; and AsTAR excels at execution pacing across long energy gaps. Finally, we demonstrate that while these advanced strategies provide critical resilience for severely constrained systems with small capacitors, devices with larger energy buffers can efficiently rely on simpler, less computationally expensive static policies.
\end{abstract}
\glsresetall

\IEEEpeerreviewmaketitle

\section{Introduction}
\label{sec:introduction}

The \ac{iot} has grown rapidly in recent years, with deployments expected to increase from around 20 billion devices today up to 40 billion by 2030~\cite{iot-analytics_2024}.
Since most devices are battery-powered, their environmental and economic impact is becoming a major concern.
Li-ion batteries contain toxic rare earth materials such as cobalt and nickel and are often improperly recycled, significantly contributing to e-waste and raising serious health concerns~\cite{zhao2021review}.
Moreover, rechargeable batteries have a limited number of charging cycles in their lifetime, leading to a large number of batteries needing replacement during long-term and large-scale deployment. This also shines a light on the economic complications that come along with battery-powered \ac{iot}, as these replacements amount to a loss of technical and financial resources that could be used elsewhere. 

In recent years, ambient \ac{iot} and batteryless \ac{iot}~\cite{3GPP_AIoT} have emerged as a response to the environmental and economic questions brought about by the rapid proliferation of \ac{iot} across all sectors of our modern society. These fields aim to significantly prolong the lifetime of battery-powered \ac{iot} devices or even eliminate the need for batteries by leveraging novel energy harvesting techniques, ultra-low-power electronics, and efficient communication protocols, paving the way for (super) capacitors as the main energy storage element. While supercapacitors typically have a higher cost per unit of energy, their extremely long cycle life and high power density make them suitable for many low-power, long-lifetime \ac{iot} applications where batteries require frequent and costly replacement~\cite{Townsend2022Comparative}. However, their lower energy density compared to batteries also introduces challenges. Unlike batteries, which maintain a relatively stable voltage during discharge, a capacitor's voltage drops proportionally to the energy it expends. This highly dynamic and non-linear behavior, coupled with the unpredictable nature of ambient energy harvesting, means that the available energy and the device's operational capabilities are constantly fluctuating. If the capacitor voltage drops below a certain critical level, it can no longer supply the \ac{iot} device with power, causing complete system failure.

Reliable operation under these dynamic energy conditions thus requires advanced energy management, only executing an application task when the capacitor is sufficiently charged.
Static operational strategies, relying on a fixed voltage threshold for task execution, are often insufficient or require hardware-specific manual calibration. Furthermore, static methods typically assume the energy consumption of a task is known in advance. In practice, applications often behave as a "black box" with dynamic energy profiles (e.g., changing energy consumption for transmitting with different payloads or transmit powers), making static prediction highly unreliable. Instead, a dynamic thresholding mechanism is desired, enabling intelligent adaptation to real-time harvested power, stored energy, and unknown task demands to optimize execution and ensure system resilience. 

\begin{figure*}[t]
\begin{center}
\includegraphics[width = 0.85\textwidth]{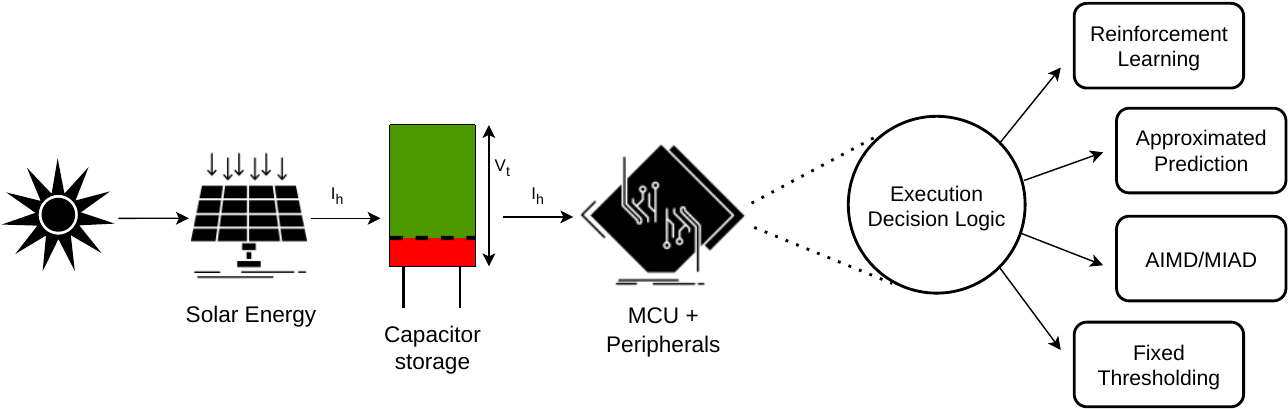}
\caption{General system overview of a batteryless ambient \ac{iot} system. Harvested solar energy gets transformed to electrical energy through a solar panel, and is stored in a supercapacitor, which powers the MCU and peripherals. The task execution logic within the MCU can be governed by multiple different approaches: Using an \ac{rl}-based approach, using approximated prediction of the workload, using AIMD/MIAD task rate adaptation, or using a fixed threshold.}
\label{fig:system_overview}
\end{center}
\end{figure*} 

To tackle this problem, we propose the general batteryless \ac{iot} system design illustrated in \Cref{fig:system_overview}. In this architecture, a solar panel converts ambient solar energy into electrical power, and the resulting harvesting current ($I_H$) is directed to a supercapacitor, which serves as the primary energy storage element. The energy stored within the capacitor then powers the \ac{mcu} and its associated peripherals. To ensure reliable operation under these highly fluctuating conditions, we investigate and compare three distinct dynamic task execution methods: AsTAR~\cite{astar2021}, which implements an \ac{aimd}/\ac{miad} approach; a novel \ac{rl}-based approach; and a novel \ac{ap}-based approach, benchmarking them against an optimized fixed threshold. Crucially, these methods are designed to operate without prior knowledge of the task's energy profile, while maintaining their adaptability across different physical capacitor sizes. Furthermore, each of these methods is based on a fundamentally different working principle, making them highly valuable for comparison. Rather than simply proving dynamic methods are superior, the core objective of this work is to expose the inherent trade-offs, such as task throughput, system recovery time, and pacing, between these approaches. We aim to identify the practical limits of dynamic thresholding and determine under which specific hardware constraints these advanced methods provide the most value compared to statically optimized baselines.

To summarize, the main contributions of this paper are:
\begin{itemize}
    \item \textbf{Hardware-Agnostic, "Black Box" Task Scheduling:} We propose two novel dynamic execution strategies capable of intelligently managing energy budgets, irrespective of capacitor size, without prior knowledge of task energy consumption: a model-free \ac{rl} agent and an on-the-fly \ac{ap} method.
    \item \textbf{Comprehensive Batteryless Simulation Framework:} We develop a physically accurate simulation environment that couples the non-linear charging dynamics of a capacitor with real-world ambient solar harvesting data and highly variable system load profiles (specifically modeling LoRa \ac{adr} transmissions). It is openly available via \href{https://github.com/SamerN97/batteryless-IoT-task-management}{https://github.com/SamerN97/batteryless-IoT-task-management}.
    \item \textbf{Evaluation of Operational Trade-offs:} We provide an extensive comparative analysis of our proposed methods against an \ac{aimd} baseline (AsTAR), a short-term oracle, and static thresholds, explicitly detailing the inherent trade-offs between task throughput, system recovery time, and execution pacing.
    \item \textbf{Identification of Practical Limits:} We highlight the boundaries of dynamic thresholding, demonstrating that these advanced methods provide the most significant value under severe hardware constraints (i.e., small capacitor sizes), while systems with a larger energy buffer can often rely on simpler, less computationally expensive static strategies.
\end{itemize}

The remainder of this paper is organized as follows: \Cref{sec:related_work} discusses existing literature on energy-aware task scheduling and energy-neutral \ac{iot} systems. \Cref{sec:problem_formulation_and_modeling} provides a comprehensive problem formulation and details the mathematical modeling of the capacitor behavior and energy dynamics. \Cref{sec:decision_logic_approaches} describes the evaluated energy-aware task execution approaches in detail. \Cref{sec:experimentation_strategy} outlines the experimental setup and the methodology employed for evaluation. \Cref{sec:results_and discussion} presents and discusses the experimental results. Finally, \Cref{sec:conclusion_and_future_work} concludes the paper and highlights future research opportunities.

\section{Related Work}
\label{sec:related_work}

The challenge of managing the dynamic energy budget in batteryless, energy-harvesting \ac{iot} devices to meet specific operational requirements spans multiple research domains, from fundamental intermittent computing architectures to advanced energy-aware scheduling algorithms. Our work intersects with several of these areas, specifically focusing on dynamic task execution without prior knowledge of energy profiles.

\subsection{Intermittent Computing and Energy Management}
A prominent strategy for managing unreliable power is intermittent computing, allowing systems to safely suspend and resume across frequent power failures. Early works like Mementos \cite{ransford2011mementos} rely on frequent state checkpointing, while subsequent models like Alpaca \cite{maeng2017alpaca} utilize task-based programming to guarantee progress without traditional checkpoints. To mitigate power loss overhead, hardware-software co-designs like Capybara \cite{colin2018capybara} introduce reconfigurable capacitor banks to prevent mid-task failures. Pushing this to the operating system level, frameworks like InK \cite{yildirim2018ink} introduce event-driven kernels that preemptively schedule reactive microtasks based on immediate energy. However, these architectures fundamentally embrace a power failure-tolerant paradigm or require specialized programming models to manage the OFF-state. In contrast, our work treats power failures as highly undesirable, focusing on proactive energy management on standard hardware to entirely avoid the OFF-state and maintain continuous capabilities.

\subsection{Energy-Aware Task Scheduling}
To proactively avoid power failures, systems must align active periods with available energy. Traditional operations use static thresholds or predictive models (e.g., \gls{ewma} \cite{kansal2007power}, solar forecasting \cite{cammarano2012proenergy}) to adjust duty cycles based on long-term averages, typically assuming deterministic execution costs \cite{moser2007adaptive}.
As recently highlighted by López et al. \cite{lopez2025foundations}, designing effective energy-aware protocols requires models that accurately reflect task-level variability and storage constraints rather than oversimplifying them. Real-time dynamic scheduling aims to address this. At the algorithmic level, AsTAR \cite{astar2021} handles fluctuating, "black box" workloads using an AIMD approach to adjust task rates. While highly effective for long-term pacing, it limits immediate energy utilization. Conversely, model-based dynamic thresholding \cite{sabovic_energy-aware_2022} recalculates the capacitor voltage to maximize immediate throughput based on instantaneous environmental changes. However, such methods still assume a deterministic workload and rely on prior task consumption profiling. Faced with non-deterministic workloads, they rely on worst-case assumptions, causing conservative, sub-optimal scheduling.
Our work bridges this gap by proposing hardware-agnostic, dynamic thresholding strategies that treat the application as a highly volatile "black box." By adapting to unknown task profiles on-the-fly, our methods aim to achieve high task throughput without offline profiling or manual calibration.

\subsection{Machine Learning for Energy Harvesting IoT}
Given ambient energy unpredictability, \ac{rl} has emerged as a promising energy management approach. Early work used tabular Q-learning to dynamically adjust wireless sensor node duty cycles based on stored energy and current harvesting rates~\cite{kosunalp2016, hsu2009}. For example, RLMan~\cite{aoudia2018} leverages linear function approximations to achieve an effective, lightweight power management policy. However, these methods, typically trained for a single, fixed hardware configuration, assume a relatively stable or predictable energy cost per active period, focusing primarily on broad duty-cycle modulation.

Recently, \ac{drl} has been applied to handle the continuous state spaces in energy harvesting environments, optimizing metrics such as throughput and latency~\cite{zhao2022}. While \ac{drl} shows promise, it generally bounds its state evaluations to environmental factors and the instantaneous state of charge, relying on fixed hardware constraints. This hardware-specific formulation is less suitable for \ac{iot} deployment scenarios with unknown task costs.

Our \ac{rl}-based approach targets "black box" applications with volatile workloads. By using a rich temporal observation space and employing robust \ac{rl} algorithms coupled with strict domain randomization of hardware parameters during training, our agent learns underlying energy dynamics instead of memorizing hardware-specific trajectories. Thus, it acts as an adaptable, hardware-agnostic scheduler capable of dynamically balancing task throughput and system recovery under unpredictable energy availability, without manual recalibration.

\section{Problem Formulation and System Model}
\label{sec:problem_formulation_and_modeling}

In batteryless, energy-neutral \ac{iot} systems, task execution decisions are tightly coupled to the energy dynamics of the storage element. For such systems, the capacitor voltage \(V(t)\) serves as the primary indicator of the available energy budget, directly influencing whether the device remains operational or risks a power failure. The system’s evolution can therefore be modeled as a \ac{mdp} in which the next state depends on the current capacitor voltage, the harvesting rate, and the decision to execute or skip a task.

At each decision point, the device must choose between two actions:
\begin{itemize}
    \item \textbf{Execute (E):} Perform a task, consuming energy.
    \item \textbf{No Execute (NE):} Go into a low-power sleep mode to accumulate energy.
\end{itemize}

 Considering a time difference $\Delta t$, the capacitor’s voltage update \(V(t+\Delta t)\), derived from its electrical model, determines whether the system stays ON or transitions to the OFF-state. Executing a task in an unfavorable energy state can result in \(V(t+\Delta t) < V_{min}\), forcing a shutdown. Conversely, skipping execution when harvesting conditions are favorable can lead to missed opportunities for useful work.

The following subsections formalize this problem in two steps:
\begin{enumerate}
    \item \emph{Capacitor Model} — We present the electrical model governing \(V(t)\) evolution under harvesting and load, and derive a closed-form update equation suitable for simulation.
    \item \emph{System Behavior} — We describe the operational modes, possible state transitions, and decision outcomes, represented as an MDP, highlighting the consequences of each action on system availability and performance.
\end{enumerate}

\subsection{Capacitor Model}
In a batteryless, energy-neutral system, the storage state is most conveniently tracked via the capacitor voltage \(V(t)\), since stored energy is described in \Cref{eq:capacitor_energy}.
Practically, \(V(t)\) determines when the system can execute work (e.g., perform a wireless transmission) and enforces safety margins to avoid power failures. To evaluate these decisions in simulation, we use a compact model that links harvesting current and load to the evolution of \(V(t)\).

\begin{equation}
    E(t)=\tfrac{1}{2}CV(t)^2
    \label{eq:capacitor_energy}
\end{equation}

\begin{figure}[!ht]
\centering
\includegraphics[width=0.25\textwidth]{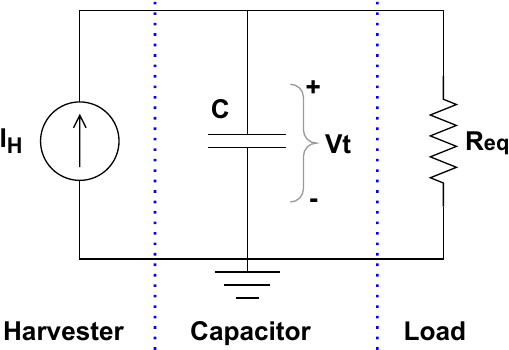}
\caption{Equivalent circuit of a batteryless \ac{iot} device with harvester current source, storage capacitor, and equivalent load \(R_\mathrm{eq}\).}
\label{fig:electrical_model}
\end{figure}

\Cref{fig:electrical_model} shows an electrical model representing an energy-neutral system containing three main components. The first component is the harvester, which is modeled as a current source $I_{H}$. It is responsible for harvesting ambient energy, such as solar or kinetic energy, and converting it to electrical energy. The second component is the capacitor \(C\), which acts as the main energy storage device of the system. The third, and final component is the load that includes all energy-consuming elements of the system, such as the CPU, radio, and inherent capacitor leakage. They are collectively modeled as an equivalent resistance $R_{eq}(t)$ at time $t$.

Applying \ac{kcl} at the storage node with Ohm’s law and the capacitor \(i\)–\(v\) relation results in:
\begin{equation}
    I_\mathrm{H}(t) = C\,\frac{dV(t)}{dt} + \frac{V(t)}{R_\mathrm{eq}(t)}.
\label{eq:kcl_expression}
\end{equation}

Intuitively, the \ac{kcl} balance \(C\,\frac{dV}{dt}=I_\mathrm{H}(t) -\frac{V(t)}{R_\mathrm{eq}(t)}\) shows that the capacitor charges when the harvester current $I_\mathrm{H}(t)$ exceeds the load current,
discharges when the load current exceeds the harvester current, and is in steady state when equality holds.

Rearranging \Cref{eq:kcl_expression} results in the following first-order linear differential equation:
\begin{equation}
    \frac{dV(t)}{dt} + \frac{1}{R_\mathrm{eq}(t) \, C}\,V(t) = \frac{I_\mathrm{H}(t)}{C}.
\end{equation}

For simulation, we assume \(I_\mathrm{H}(t)\) and \(R_\mathrm{eq}(t)\) to be constant within each step \(\Delta t\), thus dropping the function notation. Solving the first-order differential equation on \([t,\,t+\Delta t]\) then yields the exact RC step response:
\begin{equation}
    V(t+\Delta t) = I_\mathrm{H} \, R_\mathrm{eq}\!\left(1 - e^{-\frac{\Delta t}{\text{$R_{eq}$} C}}\right) + V(t)\, e^{-\frac{\Delta t}{\text{$R_{eq}$} C}}.
    \label{eq:voltage_prediction}
\end{equation}

This closed-form update is numerically stable for any \(\Delta t > 0\) and avoids per-step truncation errors introduced by numerical approximations. However, it should be noted that the physical accuracy of the prediction fundamentally relies on choosing a decision interval $\Delta t$ that is sufficiently small to ensure the piecewise-constant assumption for the harvesting current $I_\mathrm{H}$ and equivalent load $R_\mathrm{eq}$ remains valid. By mapping the continuous-time update to a discrete-time sequence where the next state $V_{t+1}$ corresponds to $V(t+\Delta t)$, this equation enables direct prediction of the storage voltage for any decision interval $\Delta t$, harvesting current $I_\mathrm{H}$, and equivalent load $R_\mathrm{eq}$. In the context of system control, \(R_\mathrm{eq}\) depends on the chosen action: a higher load during task execution (E) and a lower load during sleep (NE). Furthermore, the capacitor leakage can also be simulated as part of the equivalent load. By comparing \(V_{t+1}\) to the minimum operating voltage \(V_{min}\), the system can decide whether it will remain operational or risk a power failure (OFF) state. This link between the electrical model and decision-making forms the basis for representing the system as an \ac{mdp}, where each choice directly shapes the next energy state and operational mode.

\subsection{System Model}
The main goal of the system is to optimally utilize the harvested energy to balance high task throughput with long-term system resilience, proactively minimizing the risk of power failures. To achieve this, the device follows a duty-cycled execution pattern: it resides in an ultra-low-power sleep mode between periodic wake-ups. At each wake-up, the system samples \(V(t)\) to decide whether conditions are favorable to execute a task (E), e.g., sensing, data processing or storage, transmission, reception, etc., or whether it should return to sleep mode (NE) to accumulate more energy. We assume that a task is always ready for execution, but the energy required for task completion may vary depending on the task type. The decision is therefore based on both the available energy budget and the pending task's specific energy demand.

In a dynamic energy-harvesting environment, fixed execution schedules are unreliable: executing a task without sufficient stored energy may cause the capacitor voltage to drop below $V_{min}$, forcing the device into a power failure (OFF-state). Unlike classical intermittent computing systems, where power failures are frequent and computation resumes after state restoration, our approach aims to proactively avoid OFF-state transitions altogether through energy-aware scheduling.

Residing in the OFF-state is undesirable for several reasons. First, volatile memory is lost, requiring the system to reinitialize and reload any necessary state after power-up. Second, the device cannot perform sensing, communication, or other useful work while powered off. Finally, recovering from a power failure enforces a prolonged period of mandatory downtime, as the system remains inactive until the capacitor slowly recharges past the hardware's turn-on threshold ($V_{to}$).

\begin{figure}
    \centering
    \includegraphics[width=\linewidth]{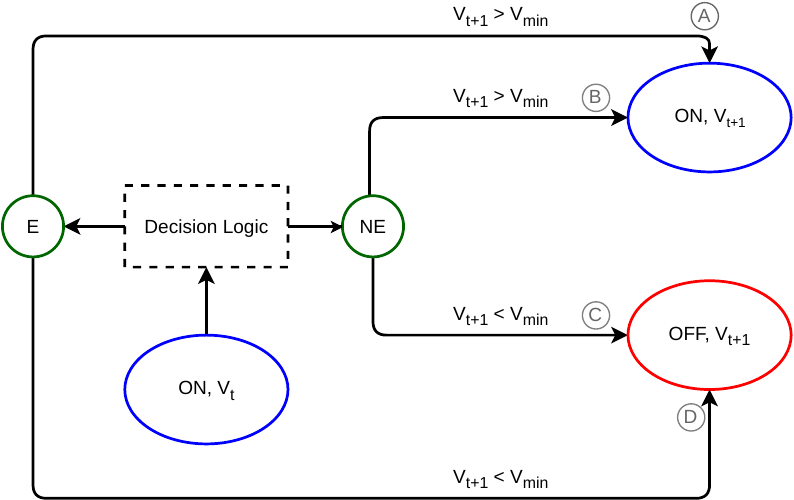}
    \caption{An overview of the possible evolutions of the system's state transitions starting from the ON state. E stands for 'execute' and represents the execution of a task, while NE stands for 'no execute' and represents the sleep mode. The ovals represent the states of the system, consisting of the device state (ON or OFF) and the capacitor voltage at time $t$ or $t+1$. \Circled{A} and \Circled{B} represent desirable state transitions that keep the device in ON-state, while \Circled{C} and \Circled{D} are undesirable transitions putting the device in OFF-state. $V_{min}$ is the minimal capacitor voltage threshold for the device to maintain its ON-state, determined by the hardware's minimal operational voltage. Note that there also exists a transition from the OFF-state to the ON-state, which is omitted from this visual overview for clarity, since it does not depend on the decision logic.}
    \label{fig:general_mdp}
\end{figure}

\Cref{fig:general_mdp} illustrates the possible system state transitions that can be determined through \Cref{eq:voltage_prediction} with \(R_\mathrm{eq}\) chosen according to the action (E or NE). Each state is defined by the device mode (ON or OFF) and the capacitor voltage at times $t$ and $t+1$:
\begin{itemize}
    \item \textbf{Safe outcomes:} The capacitor voltage at $t+1$ remains above $V_{min}$, and the device stays ON. In transition \Circled{A}, a task is successfully executed; in transition \Circled{B}, the system remains in sleep mode to conserve energy.
    \item \textbf{Failure outcomes:} The capacitor voltage at $t+1$ falls below $V_{min}$, leading to shutdown. In transition \Circled{C}, even the sleep mode is unsustainable due to insufficient harvesting current; in transition \Circled{D}, a task is attempted without sufficient stored energy, causing a power failure.
\end{itemize}

The core challenge is to determine, at each wake-up, whether executing a task will sustain system operation or cause a power failure. While the energy-harvesting environment may exhibit some degree of predictability, the energy requirements of individual tasks often vary significantly and unpredictably, making accurate decision-making more difficult. For example, the energy required for performing a wireless transmission can change depending on the network quality, the payload, and the connection time. Furthermore, the system's energy dynamics are dependent on the capacitor size. Larger capacitors result in larger energy buffers with longer charge and discharge times. The performance of the decision logic should be independent of the capacitor size. The remainder of this work aims to evaluate multiple decision logic approaches in terms of capacitor-size-agnosticism, successful task completion, OFF-state transition avoidance, and reliability.

\section{Decision Logic Approaches}
\label{sec:decision_logic_approaches}

Having modeled the system’s operational behavior as an \ac{mdp}, this section explores the decision-making policies used to govern task execution. First, we discuss the static thresholding approach, which simply triggers task execution based on a predefined capacitor voltage threshold. Then we move on to AsTAR~\cite{astar2021}, an adaptive execution mechanism based on \ac{aimd}. Finally, we introduce two novel adaptive approaches: the first one relying on an \ac{rl} agent's policy and the second one using real-time approximations of the load consumption.

\subsection{Static Thresholding Approach}

Static thresholding is the simplest policy in our evaluation. The decision logic is governed by a fixed, predefined voltage threshold, $V_{thresh}$. At each wake-up, the system compares the current capacitor voltage \(V_t\) with this fixed value. If $V_t \geq V_{thresh}$, the task will be executed (E); otherwise, the system will return to sleep mode (NE) to preserve its energy. The approach offers extreme simplicity, but its performance is highly dependent on the chosen threshold value. If a conservative approach is chosen with $V_{thresh} \gg V_{min}$, the system might miss out on execution opportunities. However, if a more ambitious approach is chosen with $V_{thresh} \approx V_{min}$, the system is at high risk of triggering power failures. While static thresholding is most effective when the energy consumption of tasks is known in advance, such deterministic energy profiles are not always available in practice.

\subsection{AsTAR Approach}

AsTAR, rather than identifying an optimal voltage threshold for task execution, aims to maintain the capacitor voltage around a specific target point, $V_{optimal}$. Consequently, its primary focus is on ensuring stable, perpetual operation rather than simply maximizing the execution frequency.

As described in Algorithm~1 of Yang et al.~\cite{astar2021}, this behavior is achieved by dynamically adjusting the task execution rate based on the current capacitor voltage, $V_t$. The adaptation logic evaluates $V_t$ against an acceptable margin ($m$) and applies different control strategies accordingly:

\begin{itemize}
    \item \textbf{Optimal Voltage ($V_{optimal} - m \le V_t \le V_{optimal} + m$):} The execution rate remains unchanged.
    \item \textbf{Low Voltage ($V_t < V_{optimal} - m$):} The execution rate is adjusted using the \ac{aimd} principle, relying on a comparison with the previously measured voltage, $V_{prev}$. If $V_t > V_{prev}$, the task rate (number of executions per day) is incrementally increased by 1. Conversely, if $V_t < V_{prev}$, the rate is divided by 2. This creates a steady increase but an aggressive decrease, prioritizing system safety during low-energy states.
    \item \textbf{High Voltage ($V_t > V_{optimal} + m$):} The system operates according to the \ac{miad} principle. This allows for a more aggressive increase in the execution rate to safely utilize excess energy, paired with a moderate decrease, which is well-suited for a secure high-voltage state.
\end{itemize}

\subsection{RL-based Approach}
\label{subsec:rl_approach}

Another approach is formulating the decision logic as a model-free \ac{rl} problem. By treating the system's energy dynamics as an \gls{mdp}, an \ac{rl} agent can learn an optimal, adaptive execution policy through continuous interaction with the environment. 
Unlike classical thresholding, the \ac{rl} agent does not calculate a definitive voltage limit. Instead, it evaluates the holistic state of the system and infers the (long-term) consequences of executing a task versus sleeping. 

For this approach, we employ \ac{ppo}~\cite{schulman2017proximal}, a highly robust actor-critic algorithm, due to its clipping mechanism. This ensures training stability and prevents catastrophic policy degradation when exposing the agent to the highly volatile, randomized capacitor bounds utilized in our training environment, which is designed with the following core \ac{rl} components:

\subsubsection*{Observation Space}
To make informed decisions, the agent receives a comprehensive, normalized state vector at each wake-up interval. To provide temporal context and allow the agent to perceive trends (e.g., a charging or discharging capacitor), the observation space includes a history window of the most recent steps for the following features:
\begin{itemize}
    \item \textbf{Capacitor Voltage ($V$):} The primary indicator of stored energy.
    \item \textbf{Harvesting Current ($I_H$):} The ambient current being generated.
    \item \textbf{Task Duration ($TD$):} The duration of the last performed task.
    \item \textbf{Time Since Failure ($TSF$):} A step counter (e.g., low-power \ac{rtc}) that tracks the elapsed time since the last power failure, functioning as a risk decay metric.
    \item \textbf{Task Parameters:} The parameters defining the energy profile of the last performed (transmission) task.
    \item \textbf{Capacitor Size ($C$):} The total capacitance of the energy storage element.
    \item \textbf{Energy level ($E$):} The last known energy level of the device, derived from $V$ and $C$ through \Cref{eq:capacitor_energy}.
\end{itemize}

\subsubsection*{Action Space}
The agent outputs a discrete action  that maps directly to the \ac{mdp} formulation described in \Cref{sec:problem_formulation_and_modeling}:
\begin{itemize}
    \item \textbf{Action (NE):} Skip execution and go into a low-power sleep state to accumulate energy.
    \item \textbf{Action (E):} Execute the task at hand.
\end{itemize}

\subsubsection*{Reward Formulation}
The \ac{rl}-based approach is flexible as the reward function can be tailored depending on the desired optimization. In this work, the reward function consists of four parts:
\begin{itemize}
    \item \textbf{\ac{ptr}:} The agent receives a positive reward for successfully completing a task execution without power failure.
    \item \textbf{\ac{pir}:} The agent receives a positive reward for choosing not to execute a task when it would have caused a power failure.
    \item \textbf{\ac{nfr}:} The agent receives a negative reward when it has experienced a power failure.
    \item \textbf{\ac{nir}:} The agent receives a negative reward for choosing not to execute a task when it would have been safe to do so.
\end{itemize}

We showcase two specific reward formulations: The first one minimizing the time between successful task executions (\ac{iti}), and the second one minimizing the total off-time. The exact reward components are formulated as follows:

\begin{table}[!h]
\centering
\renewcommand{\arraystretch}{1.5}
\begin{tabular}{@{}|l|c|c|@{}}
\toprule
\textbf{Reward} & \textbf{ITI Opt.} & \textbf{Off-Time Opt.} \\ \midrule
\textbf{PTR} & $1 + 4 \frac{TSF}{TSF_{max}}$ & $E_{norm}$ \\
\textbf{PIR} & $1$ & $1$ \\
\textbf{NFR} & $\max(-0.1 N_{off}, -1000)$ & $\max(-0.1 N_{off}, -1000)$ \\
\textbf{NIR} & $-0.5 (1 + 4 E_{norm} \, I_{norm})$ & $-0.5 E_{norm} \, I_{norm}$ \\ \bottomrule
\end{tabular}
\end{table}


The main objective of the \ac{iti} Optimization is to enforce consistent task execution by minimizing the variance and bounding the maximum time between successful tasks. To achieve this, the \ac{ptr} is scaled by the device's survival streak ($TSF$ with $TSF_{max} = 1000$). Its design teaches the agent that maintaining continuous, stable uptime is a necessary prerequisite for frequent execution. However, to prevent the agent from artificially inflating its $TSF$ by simply remaining idle, the penalty for inaction (\ac{nir}) is aggressively amplified when both the stored energy and ambient harvesting conditions are high. This forces the agent to actively capitalize on available energy rather than hoarding it, thereby closing the gaps between executions.

The Off-Time Optimization, on the other hand, is designed with a more conservative, survival-first objective: minimizing the absolute time the system spends in a power failure state. Here, the agent's rewards are tightly coupled strictly to the available energy budget. By scaling the \ac{ptr} directly with the normalized capacitor energy level $E_{norm}$, the agent is incentivized to execute tasks primarily when the capacitor is well-charged, inherently buffering the system against sudden energy depletion. While the NIR still discourages the agent from wasting peak harvesting opportunities, the absence of a $TSF$ multiplier means the agent operates without the aggressive pressure to maintain an execution rhythm, resulting in a safer, highly robust policy.

To establish a baseline of system safety, both formulations share a static reward of 1 (PIR) for correctly choosing to sleep when executing a task would have caused a power failure. Furthermore, both utilize a dynamic Negative Failure Reward (NFR) that scales proportionally with the duration of the off-state ($N_{off}$). This directly aligns the mathematical penalty with the exact physical metric the system aims to minimize, ensuring the agent learns to avoid prolonged periods of failure.

\subsubsection*{Domain Randomization for Generalization}
To ensure hardware-agnostic performance, it is important to expose the agent to variability in capacitor sizes during training. This is achieved through domain randomization of the simulated hardware. The capacitor size ($C$) is randomly mutated across different training episodes. This forces the policy to learn the underlying physics of the capacitor's charge and discharge behavior rather than memorizing a fixed trajectory for a specific hardware configuration.

\subsubsection{Energy and Memory Overhead}

While the \ac{rl} agent provides a highly adaptive policy, its execution introduces an inherent computational overhead that must be accounted for within the system's energy budget.
Because the \ac{ppo} model is trained offline, the \ac{mcu} is strictly responsible for inference. However, the process of extracting the observation state, loading the network parameters, and executing the mathematical forward pass of the neural network requires active processing time, drawing a higher current than the device's baseline sleep mode. Consequently, the equivalent electrical load of the agent ($R_{agent}$) and its required execution time ($t_{agent}$) must be explicitly factored into the simulation.
Furthermore, deploying the \ac{rl} agent introduces specific memory overheads for the \ac{mcu}. The compiled actor-network weights and biases must be persistently stored in Flash memory, while runtime RAM is required to maintain the observation history array, which provides the agent with essential temporal context.

\subsection{Approximated Prediction Approach} \label{sec:model-based}

Traditional model-based approaches, such as the one presented in~\cite{sabovic_energy-aware_2022}, rely on prior knowledge of a task's specific power consumption. Because we treat the application as a "black box" with unknown energy demands, directly applying these models is unfeasible. However, this limitation can be overcome by dynamically approximating the task's power consumption based on measurable system parameters.

Looking back at \Cref{eq:voltage_prediction} for simulating the capacitor behavior, the only unknown variable required to predict the future voltage $V(t+\Delta t)$ is the equivalent load $R_\mathrm{eq}$. Assuming that a task's power profile does not change drastically over a short period, we can approximate the current task's consumption by deriving $R_\mathrm{eq}$ from the execution of the previous task. By measuring the capacitor voltage immediately before ($V_{t-1}$) and after ($V_t$) a task completes, we capture the necessary information to infer $R_\mathrm{eq}$. 

Because $R_{eq}$ is present both inside and outside the exponential term, rearranging \Cref{eq:voltage_prediction} results in an implicit relationship that cannot be solved for $R_{eq}$ directly. Therefore, we formulate this as a root-finding problem to numerically approximate $R_{eq}$ using the Newton-Raphson method, defining our objective function $f(R_{eq}) = 0$ as follows:

\begin{equation}
f(R_{eq}) = R_{eq} + \frac{t}{C \ln\left( \frac{V_t - I_H R_{eq}}{V_{t-1} - I_H R_{eq}} \right)} = 0
\label{eq:Req_objective}
\end{equation}

While the Newton-Raphson solver provides an accurate estimate for $R_{eq}$, each approximation introduces a computational overhead, which translates to an added energy cost. To minimize this impact on the energy budget, we introduce a predict-and-verify execution strategy coupled with a dynamic recalibration procedure, as outlined in Algorithm~\ref{alg:recalibration}.

The system uses the most recently approximated $R_{eq}$ to predict the future capacitor voltage $V_{predict}$ that would result from executing the pending task. The execution decision is based directly on this prediction: if $V_{predict}$ safely exceeds a padded minimum operating threshold $V_{min\_padded}$, the system executes the task. To account for estimation inaccuracies, this padding scales proportionally with the approximated task current $I_{task}$, such that $V_{min\_padded} = V_{min} + k \cdot I_{task}$, where the scaling constant is experimentally determined to be $k = 1.5\ \text{V/A}$. This ensures a larger safety buffer for heavier, more power-hungry executions. If the predicted voltage is too low, the system skips the execution and returns to sleep mode to harvest more energy.

Following a task execution, the system enters a verification phase to determine if a recalibration is needed. By introducing an error threshold $E_{thresh}$, which is dynamically scaled to the currently estimated task consumption ($10 \cdot V_{supply}/R_{eq}$), the system evaluates the accuracy of the cached $R_{eq}$. A full recalibration, consisting of a new Newton-Raphson approximation and an update to $E_{thresh}$, is triggered on only two occasions. The first case is when the system undergoes a power failure as a result of a task execution. This absolute failure implies that the current $R_{eq}$ was severely underestimated and must be recalculated upon reboot. The second case is when the absolute difference between the predicted voltage $V_{predict}$ and the actual measured voltage $V_{actual}$ after task execution exceeds the error threshold $E_{thresh}$. This indicates that the current $R_{eq}$ is drifting from the system's estimate due to a shifting power profile, meaning the model must be recalculated to ensure future reliability.

\begin{algorithm}[t]
\caption{Approximated Prediction Calibration Procedure}
\label{alg:recalibration}
\small
\KwIn{Current voltage $V_t$, Harvested current $I_H$, Hardware minimum voltage $V_{min}$, Supply voltage $V_{supply}$}
$\text{CalibrateFlag} \gets \text{True}$\;
$E_{thresh} \gets \text{Initial Error Threshold}$\;

\BlankLine
\ForEach{wake-up interval}{
    \eIf{$\text{CalibrateFlag} == \text{True}$}{
        \eIf{$V_t > V_{thresh}$}{
            $V_{t-1} \gets V_t$\;
            Execute Task\; 
            Measure $V_t$\;
            $R_{eq} \gets \text{NewtonRaphson}(V_{t-1}, V_t, I_H, t)$\;
            $E_{thresh} \gets 10 \times (V_{supply} / R_{eq})$\; 
            $\text{CalibrateFlag} \gets \text{False}$\;
        }{
            Sleep()\;
        }
    }{
        $V_{predict} \gets \text{PredictVoltage}(V_t, I_H, R_{eq})$\;
        
        \eIf{$V_{predict} > V_{min\_padded}$}{
            Execute Task()\;
            $V_{actual} \gets \text{MeasureVoltage}()$\;
            
            \uIf{$V_{actual} \leq V_{min}$}{
                $\text{CalibrateFlag} \gets \text{True}$\; 
            }
            \uElseIf{$|V_{actual} - V_{predict}| > E_{thresh}$}{
                $\text{CalibrateFlag} \gets \text{True}$\; 
            }
        }{
            Skip Task and Sleep()\;
        }
    }
}
\end{algorithm}

\subsubsection{Energy and Memory Overhead}
While the \gls{ap} approach provides a lightweight mechanism for dynamic thresholding, it introduces specific computational and memory overheads. The energy costs occur (1) during prediction, where evaluating the exponential update equation with a cached \gls{req} requires brief \gls{mcu} active time, and (2) during calibration, where it runs a higher cost, iteration-dependent Newton-Raphson solver. To avoid power failures, this should be explicitly modeled in the simulation, and the maximum number of operations should be bound (max 100 iterations used in this work). Regarding memory, non-volatile Flash is required to store the algorithmic solver logic and mathematical libraries (such as floating-point exponentials and logarithms). Runtime RAM requirements remain minimal, needing only enough space to cache recent $R_{eq}$ values, pre- and post-task voltage measurements ($V_{t-1}, V_t$), and the dynamic error threshold ($E_{thresh}$).

\subsection{Short-Term Oracle}
Finally, for the purpose of evaluation, we also introduce a \ac{st} Oracle baseline. The \ac{st} Oracle serves purely as a comparative approach and cannot be implemented in real-world scenarios. It operates with perfect short-term knowledge of the outcome of an execution decision; therefore, it always executes a task as long as that specific execution will not immediately cause the device's capacitor voltage to drop below $V_{min}$ and trigger a power failure. Note that this approach is not globally optimal, as it does not look beyond a single task execution.

\section{Experimentation Strategy and Parameters}
\label{sec:experimentation_strategy}

To evaluate the performance of our proposed adaptive decision logic approaches, we developed a comprehensive simulation framework. This framework accurately models the hardware characteristics, task execution profiles, and energy harvesting dynamics of a realistic batteryless \ac{iot} device. The experiments are designed to test the robustness, energy efficiency, and hardware-agnostic capabilities of each scheduling policy under changing environmental conditions.

\subsection{System and Task Modeling}
Our simulated batteryless \ac{iot} node is modeled as a fully tasked sensing and communication device. It simulates a central STM32L4 \ac{mcu}~\cite{stmicro_stm32l412} interfacing with an SHT30 temperature and humidity sensor~\cite{sensirion2022sht3x}, and an SX1262 LoRa communication module~\cite{semtech_sx1262} capable of variable payloads, spreading factors, and transmit power levels controlled through an \ac{adr} mechanism, dynamically adapting the transmission's power profile to the current wireless channel conditions~\cite{ttn_adr_docs}.

For our experiments, the base decision interval for the system wake-ups was set to $t_i = 30$~seconds. At each wake-up, the system always measures $V_t$ (ADC read-out) and $I_H$ (Coulomb counter read-out) and performs a temperature measurement. Subsequently, the decision logic is performed before committing to either executing a LoRa transmission or returning to deep sleep. The decision logic, therefore, only applies to the transmission task as this is the main energy consumer within the system.

To ensure the physical accuracy of the \ac{mdp} and the capacitor update equations detailed in \Cref{sec:problem_formulation_and_modeling}, the power profiles of the individual hardware components were explicitly defined. The operational voltage bounds were set to a maximum capacitor voltage $V_{{max}} = 5.5$~V, a hardware turn-on threshold $V_{{to}} = 2.3$~V, and a minimum operational threshold $V_{{min}} = 1.8$~V. Dropping below $V_{{min}}$ immediately triggers a power failure (OFF-state).  

The equivalent load resistance $R_{{eq}}$ for the system dynamically shifts based on the energy consumption of the task.

\subsubsection{Fixed Energy Consumption Profiles}
The baseline operations of the device exhibit deterministic execution times and current draws. These are based on the electrical characteristics defined in the respective component datasheets~\cite{stmicro_stm32l412, sensirion2022sht3x, semtech_sx1262} operating at a 3.3~V supply. For computational tasks, the \ac{mcu} clock frequency is assumed to be 1~MHz. The \ac{rl} policy is implemented with a total of 20,163 parameters. At a clock frequency of 1~MHz, the inference process requires approximately 44,000 clock cycles for multiply-accumulate operations and non-linear activations, resulting in an estimated 53~ms of active processing time when considering 20\% control and data overhead. The specific energy consumption for each fixed activity is outlined in \Cref{tab:fixed_energy_consumption}.

{\small
\begin{table}[t]
    \centering
    \caption{Fixed Energy Consumption Per Task ($V_{supply}$ = 3.3V)}
    \begin{tabular}{l l l l}
        \toprule
        \textbf{Component} & \textbf{\begin{tabular}[c]{@{}c@{}}Current\\(mA)\end{tabular}} & \textbf{\begin{tabular}[c]{@{}c@{}}Duration\\(s)\end{tabular}} & \textbf{\begin{tabular}[c]{@{}c@{}}Energy\\(mJ)\end{tabular}} \\
        \midrule
        Sleep (standby)   & 0.00065  & \textendash & \textendash \\
        Sense temp.  & 0.691 &  0.00533  & 0.0122 \\
        Read $V_t$ & 0.311 & 0.00005 & 0.00005 \\
        Read $I_H$ & 0.091 & 0.00023 & 0.00007 \\
        Read \ac{rl} model & 0.091 & 0.00504 & 0.00151 \\
        Run \ac{rl} model & 0.091 & 0.053 & 0.0159 \\
        Run \ac{ap} sim. & 0.091 & 0.0003 & 0.00009 \\
        \midrule
        \multicolumn{4}{l}{\footnotesize
        \textit{Note: Capacitor leakage energy consumption not included.}} \\
        \bottomrule
    \end{tabular}
\label{tab:fixed_energy_consumption}
\end{table}
}

\subsubsection{Variable Energy Consumption Profiles}

Unlike the baseline operations, several critical components of the system exhibit highly variable energy demands depending on the environmental state, algorithmic behavior, and hardware configuration. These dynamics are explicitly modeled to assess the adaptability of the different scheduling approaches:

\begin{itemize}
    \item \textbf{LoRa Transmission (\ac{adr}):} The communication task's energy profile is highly fluctuating. To simulate real-world dynamic channel conditions, such as log-normal shadow fading caused by environmental changes, we generate an RSSI trace based on a normal distribution. The trace uses a baseline of -110~dBm and is bounded between -135~dBm and -90~dBm. Based on the instantaneous RSSI, the \ac{adr} mechanism adjusts the Spreading Factor ($SF \in [7, 12]$) and transmit current (15~mA to 118~mA) according to Semtech specifications~\cite{semtech_sx1262}, mapped in \Cref{tab:adr_mapping}. To simulate varying application demands, the payload ($PL$) size is randomized between 20 and 255 bytes on a daily basis. The reception window is fixed at 50~ms with a 12~mA current draw.

    To accurately simulate transmission energy, the total air-time ($T_{air}$) is calculated based on $SF$ and $PL$. Following standard LoRa specifications, $T_{air}$ is defined as:
    $$T_{air} = \left(n_{preamble} + 4.25 + n_{payload}\right) \cdot \frac{2^{SF}}{BW}$$
    where the bandwidth $BW = 125,000$~Hz, the programmed preamble $n_{preamble} = 8$, and the payload symbol count $n_{payload}$ is calculated as:
    $$n_{payload} = 8 + \max\left(\left\lceil \frac{8PL - 4SF + 44}{4(SF - 2DE)} \right\rceil \cdot CR, 0\right)$$
    Here, $CR = 5$ (for a standard 4/5 coding rate) and the Low Data Rate Optimization flag $DE = 1$ if $SF \ge 11$ (and $0$ otherwise). The total transmission energy is then obtained from the product of $T_{air}$, the 3.3~V supply voltage, and the dynamic \ac{adr} transmit current.
    
    \item \textbf{\ac{ap} Calibration Procedure:} When the \ac{ap} approach triggers a calibration, it relies on a Newton-Raphson solver to estimate the equivalent load resistance. Because the execution time depends on the number of iterations required to converge, its energy consumption is variable. Operating at 1~MHz, each iteration requires approximately 3~ms at an active current draw of 0.091~mA, with the solver capped at a maximum of 100 iterations to prevent infinite loops.
    
    \item \textbf{Capacitor Leakage:} To ensure hardware-agnostic evaluation, the simulation utilizes varying capacitor sizes. Because inherent leakage current scales with the physical size of the capacitor, a static leakage value is insufficient. Based on the datasheet specifications for the 5.5~V DGH capacitor series~\cite{cde_dgh_datasheet}, we derived a linear regression model to dynamically calculate the leakage current $I_{leakage}$ (in $\mu$A) based on the capacitance $C$ (in Farads): $I_{leakage}=4.7442\cdot C+4.9302$.
\end{itemize}

{\small
\begin{table}[t]
    \centering
    \caption{LoRa \ac{adr} Parameter Mapping Based on RSSI ($V_{supply}$ = 3.3V)}
    \label{tab:adr_mapping}
    \renewcommand{\arraystretch}{1.2}
    \begin{tabular}{@{}lccc@{}}
        \toprule
        \textbf{RSSI Range (dBm)} & \textbf{SF} & \textbf{TX Power (dBm)} & \textbf{TX Current (mA)} \\ \midrule
        $\leq -125$ & 12 & +22 & 118 \\
        $(-125, -120]$ & 10 & +17 & 58 \\
        $(-120, -110]$ & 9 & +14 & 45 \\
        $(-110, -100]$ & 8 & +14 & 25.5 \\
        $> -100$ & 7 & +10 & 15 \\ \bottomrule
    \end{tabular}
\end{table}
}

\subsection{Environmental Data and Preprocessing}

\begin{figure}[t]
\centering
\includegraphics[width = 0.48\textwidth]{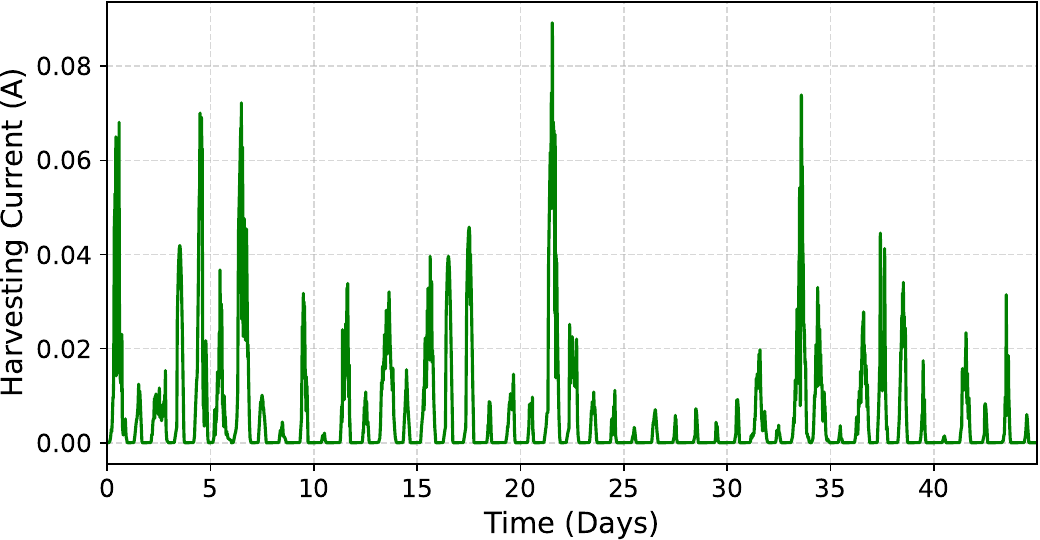}
\caption{Overview of the solar harvesting validation data used in the experiments with an augmented sample frequency of 30 seconds.}
\label{fig:solar_data}
\end{figure}

To simulate realistic and highly dynamic ambient energy availability, we used a dataset containing solar irradiance data collected from September 2022 until August 2023 in Hverager\dh{}i, Iceland. We then converted the irradiance data to harvesting current by considering a solar panel with size 60.1x41.3~mm and efficiency of 18.5\% (in line with commercially available panels), approximated cosine loss of 25\%, \ac{pmic} efficiency of 90\%, and supply voltage of 3.3~V, resulting in a conversion factor of 0.000093942. Since the original dataset was sampled at 15-minute intervals, we applied linear interpolation to upsample the data, generating 29 intermediate points between each raw sample. This provided a high-resolution harvesting current ($I_H$) profile that aligns with our 30-second system decision interval. Furthermore, the data was cleaned of "midnight sun" days, chunked into 72-hour segments and randomly shuffled to eliminate seasonal pattern influences. Finally, the data was split into a training set (80\%) and a validation set (20\%). Comprising 45 days of solar harvesting data, the validation set serves as the final evaluation benchmark for all approaches. As illustrated in \Cref{fig:solar_data}, this dataset captures a diverse range of weather conditions and daylight hours.

\subsection{RL Training and Inference Configuration}

The \ac{rl} agent was trained using the \ac{ppo} implementation from the StableBaselines3 framework~\cite{stable-baselines3}. To ensure hardware-agnostic performance, the environment employs domain randomization, i.e., at the start of every training episode, the capacitor size $C$ is uniformly sampled from a discrete set of values $\zeta$ ranging from 0.5~F to 10~F, in increments of 0.5~F, i.e., $\zeta = \{0.5, 1.0, \dots, 10.0\}$~F.

 To provide the agent with temporal context and allow it to perceive charge/discharge trajectories, the observation space is stacked with a history window of the $k=10$ most recent timesteps. 

The training and environment parameters were configured as follows:
\begin{itemize}
    \item \textbf{Algorithm:} \ac{ppo} with default \textit{MlpPolicy} architecture and hyperparameters
    \item \textbf{Discount factor ($\gamma$):} 0.99
    \item \textbf{Training Episode Length:} 8,640 steps (equivalent to 3 physical days)
    \item \textbf{Total Training Steps:} 1,000,000
\end{itemize}

A critical design choice is decoupling the physical simulation clock from the agent's steps during power failures. When the device shuts down ($V_t < V_{{min}}$), the entire physical recovery period required to reach the turn-on threshold ($V_{{to}}$) is perceived by the agent as a single, heavily penalized step. This accurately mirrors physical deployment: since the \ac{mcu} is powered down during a failure, the agent cannot actively experience the passage of time. Furthermore, strict temporal boundaries are maintained: if the physical simulation clock reaches the maximum episode length during this OFF-state, the episode truncates immediately.

For the evaluation phase, the inference strategy shifts from episodic training to continuous deployment. The episode length is extended to match the exact size of the validation dataset (45 days, or 129,600 steps).

\section{Results and Discussion}
\label{sec:results_and discussion}
In this section, we present and analyze the experimental results from our simulated batteryless \ac{iot} environment. We compare the performance of the different decision logic approaches across multiple criteria. The evaluation focuses on exposing the inherent trade-offs between task throughput, system reliability, and hardware agnosticism under highly variable solar energy harvesting conditions.
\subsection{Evaluation Setup and Metrics }

To comprehensively evaluate the robustness and adaptability of each decision logic approach, the simulation was run across the full validation dataset of solar harvesting traces. To test hardware-agnosticism, each approach was evaluated for all capacitor sizes in $\zeta$. 

Furthermore, in addition to an optimized static thresholding approach, we introduced two more threshold levels to represent opposite ends of the operational spectrum. The first, Static (1.9~V), is an aggressive threshold set at 0.1~V above the hardware's minimum operating voltage ($V_{{min}} = 1.8$~V), designed to prioritize immediate task execution. The second, Static (3.45~V), uses the optimal threshold level when using a 0.5~F capacitor, and represents a highly conservative approach that prioritizes system survival by demanding a large energy buffer before allowing any task execution.

The performance of each policy was assessed using five metrics:

\begin{itemize}
    \item \textbf{Mean daily successful executions:} The average number of tasks successfully completed within a 24-hour period without causing the capacitor voltage to drop below the minimum operational threshold ($V_t < V_{{min}}$).
    \item \textbf{Median time between off-states (survival time):} The median continuous duration the system maintains an operational state ($V_t \ge V_{{min}}$) before experiencing a power failure.
    \item \textbf{Mean \acrfull{iti}:} The average elapsed time between two consecutive successful task executions across the entire evaluation period.
    \item \textbf{Median daily maximum \acrfull{iti}:} The median of the single longest interval between consecutive successful executions recorded for each 24-hour period. This metric specifically captures the system's ability to pace its energy consumption across prolonged harvesting gaps (e.g., nighttime).
    \item \textbf{Median continuous off-state duration (recovery time):} The median time the system remains non-operational after a power failure before the capacitor accumulates sufficient energy to reach the hardware turn-on threshold ($V_t \ge V_{{to}}$).
\end{itemize}

\begin{figure}[t]
\centering
\includegraphics[width = 0.48\textwidth]{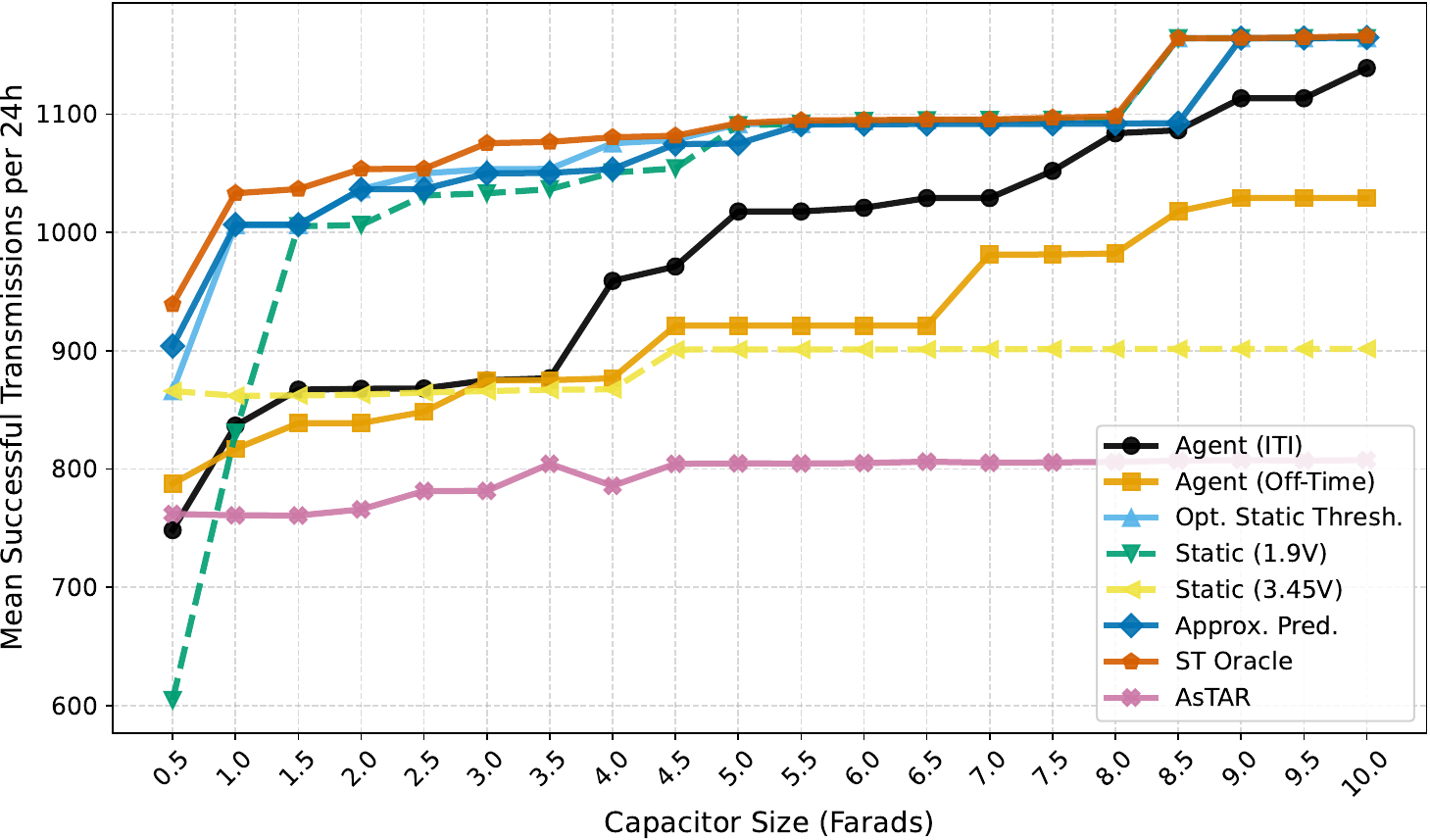}
\caption{Mean daily successful executions by approach for capacitor sizes 0.5-10 F (0.5 F steps).}
\label{fig:daily_success}
\end{figure} 

\begin{figure}[t]
\centering
\includegraphics[width = 0.48\textwidth]{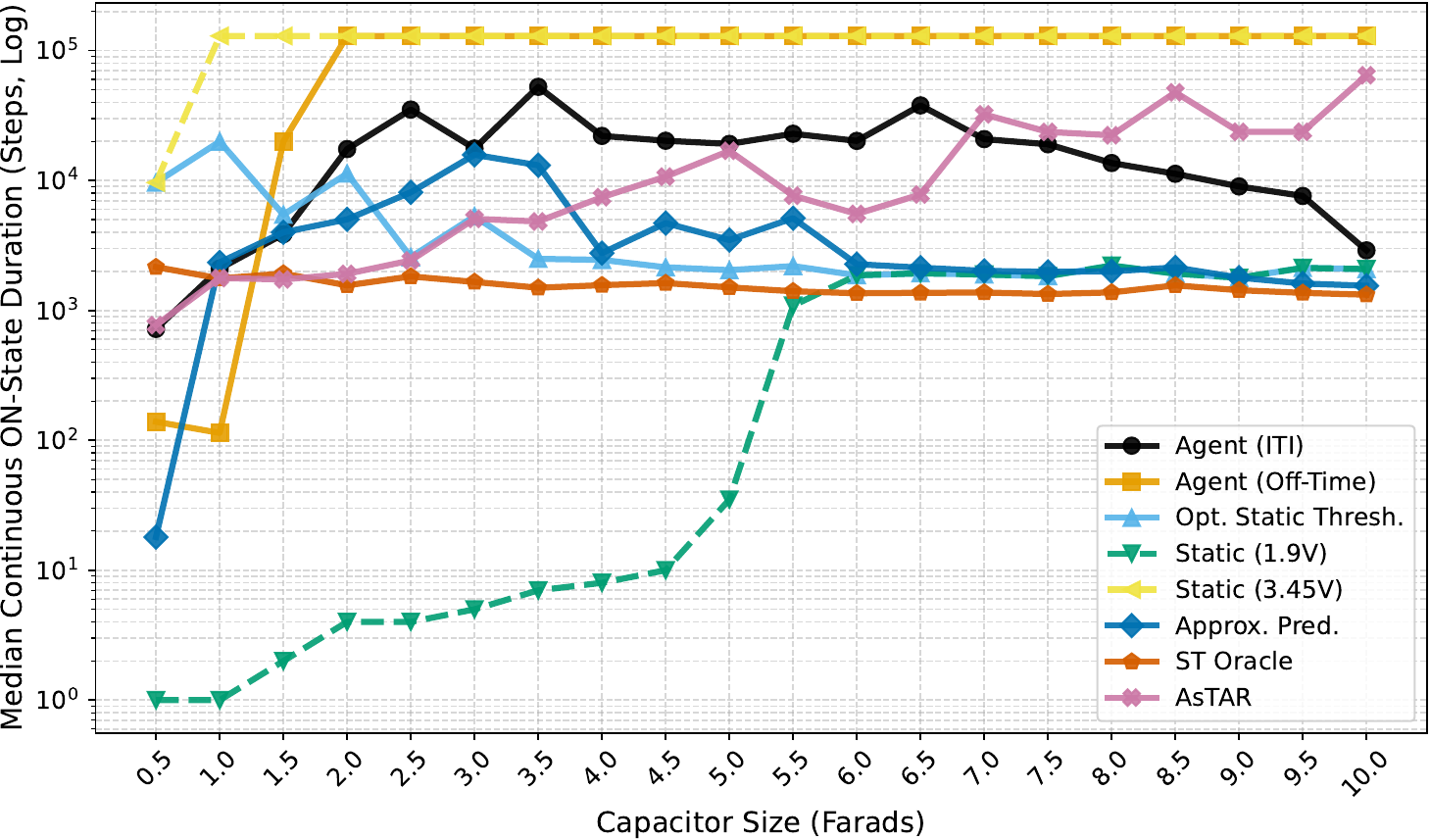}
\caption{Median time between off-states by approach for capacitor sizes 0.5-10 F (0.5 F steps).}
\label{fig:time_between_off_states}
\end{figure}

\subsection{Execution rate vs. Reliability}
A central challenge in batteryless \ac{iot} is balancing the desire for a high execution rate with the necessity of system reliability. This trade-off becomes clear when comparing the mean daily executions against the median time between off-states.
The results demonstrate a "Boom and Bust" principle among the more greedy methods. \Cref{fig:daily_success} shows approaches such as the \ac{st} Oracle, Optimal Static Threshold, Static (1.9~V), and \ac{ap} achieving the highest number of daily successes, already exceeding 1000 transmissions per 24 hours starting from the 1-1.5 F capacitor size. It is important to explicitly highlight the performance of the \ac{ap} approach here: while the \ac{st} Oracle relies on perfect near-future knowledge and the Optimal Static Threshold requires prior optimization, \ac{ap} achieves near-Oracle throughput without requiring any prior knowledge of the task or capacitor size. It dynamically calculates on-the-fly what the optimal static method must find through prior trial and error. However, \Cref{fig:time_between_off_states}, indicates that an aggressive execution profile comes at a cost to reliability. These greedy methods exhibit significantly lower median times between off-states, indicating more frequent power failures. While these approaches maximize the total number of data transmissions, the resulting increase in off-time implies a loss in continuous baseline sensing capabilities (e.g. temperature monitoring), as the device is entirely inactive during these blackout periods.

\subsection{Pacing}
\begin{figure}[t]
\centering
\includegraphics[width = 0.48\textwidth]{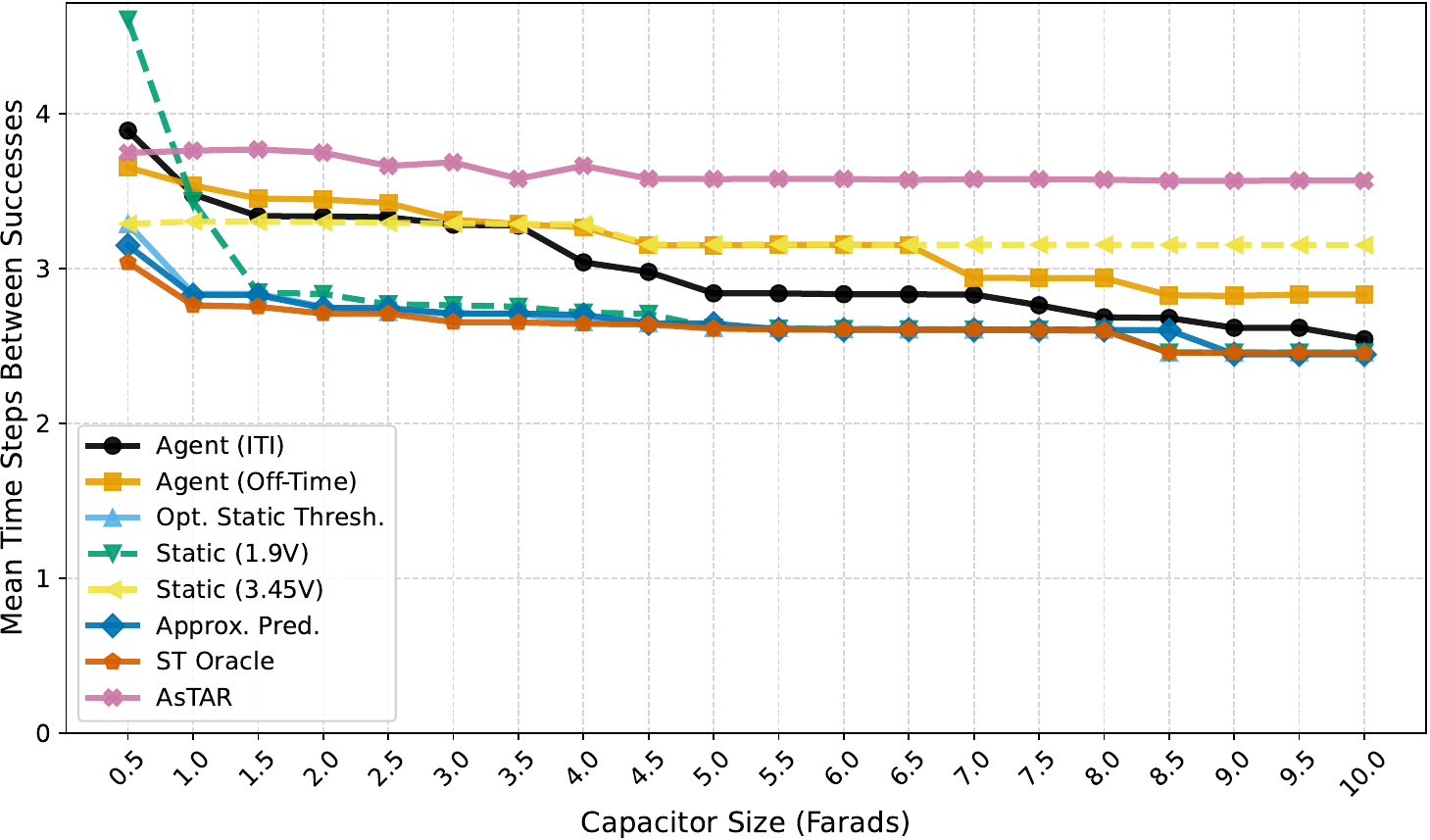}
\caption{Mean time between successful executions by approach for capacitor sizes 0.5-10 F (0.5 F steps).}
\label{fig:mean_iti}
\end{figure}

\begin{figure}[t]
\centering
\includegraphics[width = 0.48\textwidth]{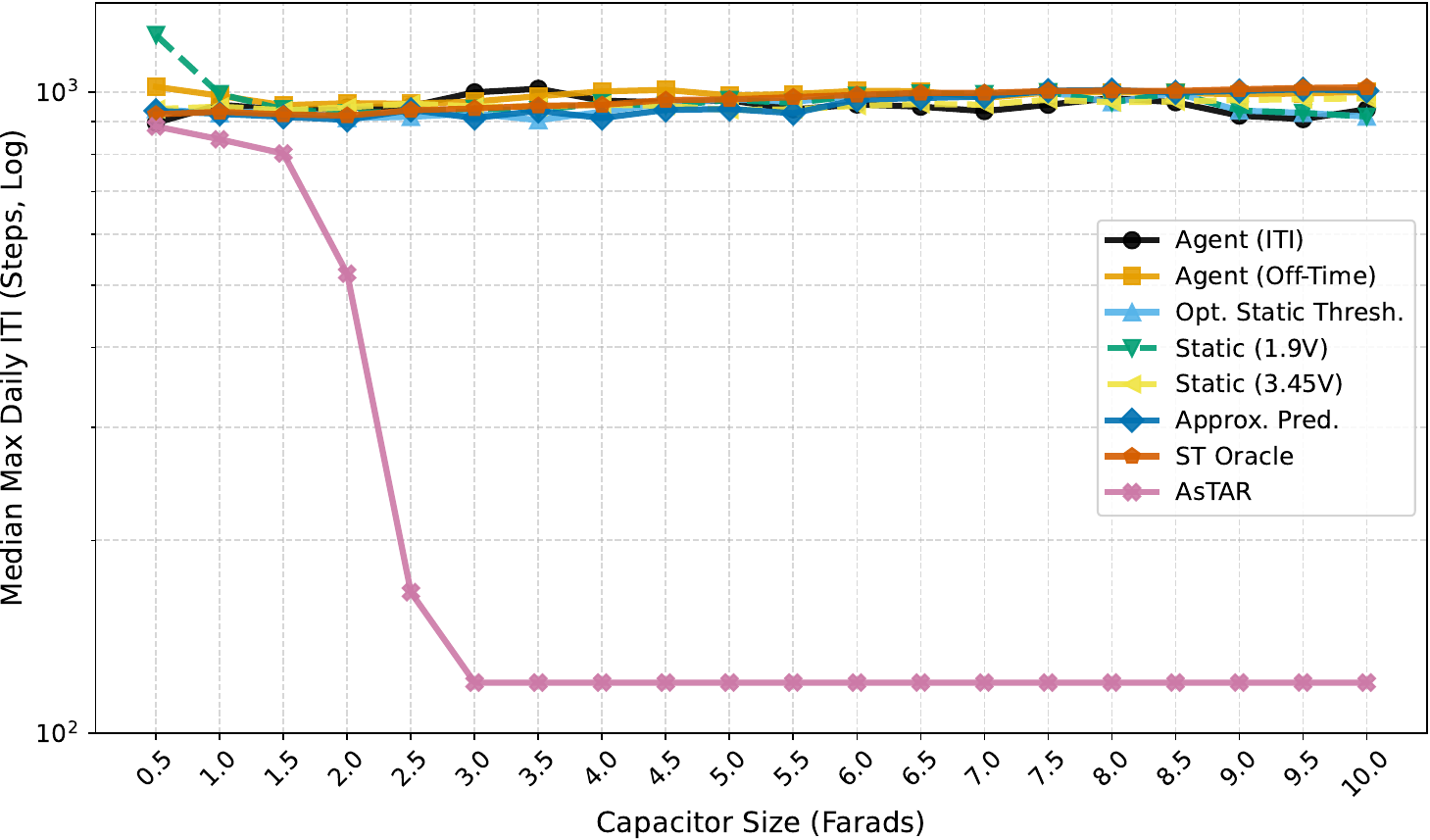}
\caption{Median of the daily maximum time between executions by approach for capacitor sizes 0.5-10 F (0.5 F steps).}
\label{fig:max_daily_iti}
\end{figure}

To understand how executions are distributed over time, we analyzed the mean \ac{iti} and the daily maximum \ac{iti}. Looking at \Cref{fig:mean_iti}, the AsTAR approach consistently maintains the highest average time between executions across all capacitor sizes, and therefore the lowest average execution rate. However, examining \Cref{fig:max_daily_iti}, the daily maximum \ac{iti} reveals AsTAR's distinct advantage.
For all other approaches, the median maximum ITI flatlines at approximately $10^3$ steps (roughly 8 hours), which directly correlates to the nighttime period where solar harvesting is zero. While AsTAR executes less frequently on average, it is specifically designed to effectively bridge this nighttime gap. For capacitor sizes larger than 2.5~F, AsTAR drastically reduces the maximum daily ITI to near $10^2$ steps (around 1 hour). It is the only evaluated approach that successfully paces its energy consumption to retain active communication capabilities throughout the night, eliminating the standard 8-hour communication blackout. In scenarios where night-time transmissions are strictly required, AsTAR therefore presents itself as the preferred approach.

\begin{figure}[t]
\centering
\includegraphics[width = 0.48\textwidth]{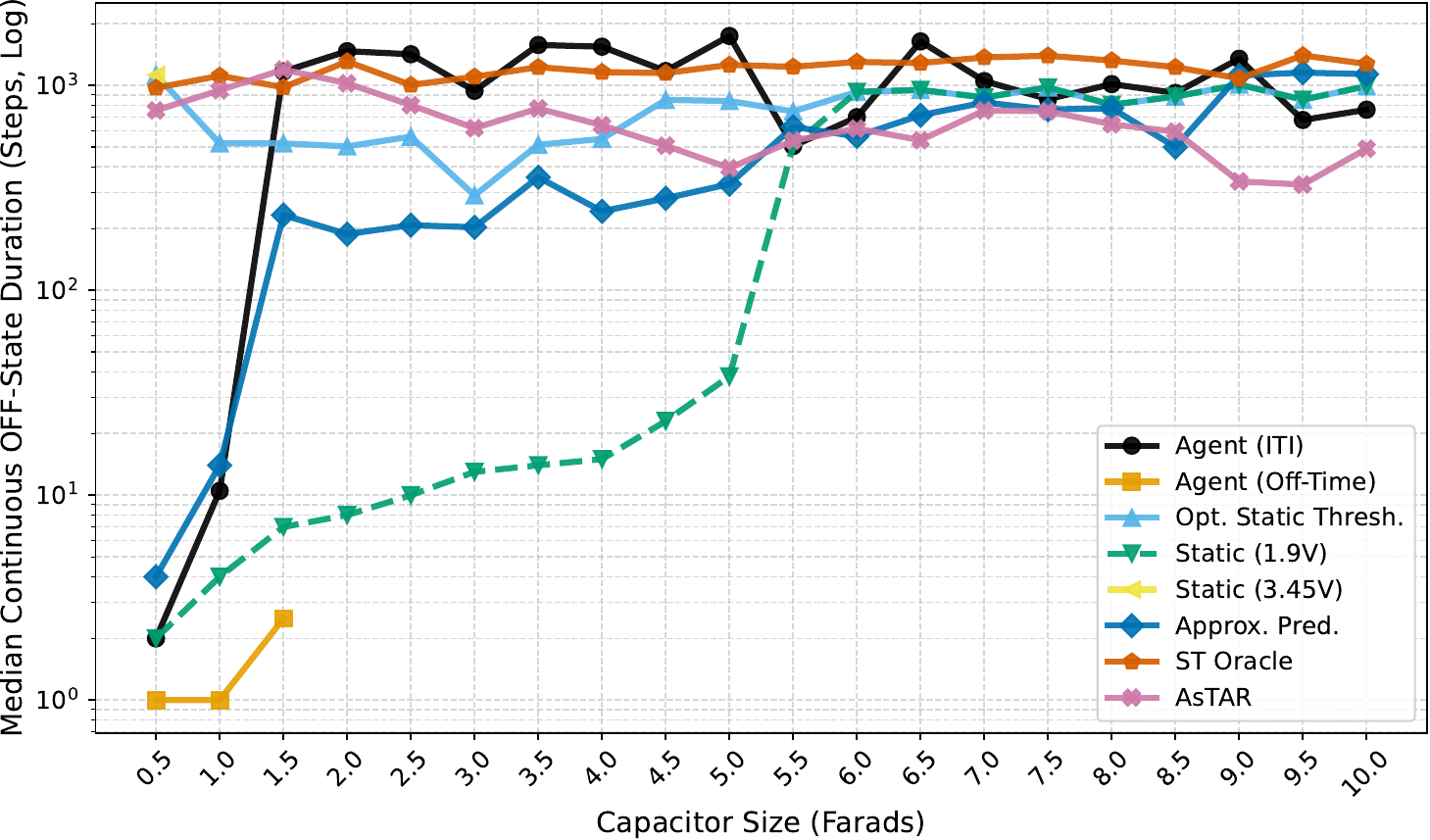}
\caption{Median of the continuous off-state duration by approach for capacitor sizes 0.5-10 F (0.5 F steps). Notice the Agent (off-time) and Static (3.45) approaches completely eliminating off-time starting from 2~F and 1~F respectively.}
\label{fig:off_state_duration}
\end{figure}

\subsection{System Resilience and Recovery Times}
System resilience is not only defined by avoiding power failures but also by how quickly a device can recover when one occurs. \Cref{fig:off_state_duration} highlights major differences in recovery behaviors. The Static (3.45~V) and the Agent (Off-Time) approaches stand out by effectively eliminating power failures completely for the vast majority of capacitor sizes, thereby maximizing total device up-time.
Interestingly, in the highly constrained, low-capacitor region (0.5 F and 1 F), the \ac{rl} Agents and the \ac{ap} approach show specific recovery adaptations. In this range, they manage to keep off-state durations remarkably low, successfully waking up and establishing operation faster than the static or Oracle approaches. Furthermore, in the 1.5~F to 6~F range, the \ac{ap} approach also manages to keep the off-state duration lower than the Optimized Static approach while generally outperforming it in terms of survival time and achieving equal performance in mean daily executions. This demonstrates that AP's dynamic recalibration not only matches the throughput of statically optimized baselines but actively protects the system better during recovery phases.

\subsection{Hardware Agnosticism}
A major limitation of static thresholding is the need for manual optimization depending on the specific hardware deployment. The experimental results show that a low threshold like Static (1.9~V) struggles to keep the device alive in the 0.5~F to 5.5~F range, leading to massive drops in survival time (\Cref{fig:time_between_off_states}). However, the survival time for Static (1.9~V) significantly increases at 5.5~F. Because the stored energy scales with capacitance (c.f. \Cref{eq:capacitor_energy}), 1.9~V now represents enough residual energy to prevent a drop below $V_{{min}}$. Conversely, picking a highly conservative threshold like Static (3.45~V) ensures device survival across the board, but leaves large amounts of harvested energy unutilized, resulting in missed execution opportunities and a lower daily success rate (\Cref{fig:daily_success}).
The adaptive methods, particularly the RL agents and AP approach, demonstrate strong hardware agnosticism. They dynamically adjust their execution logic without requiring manual recalibration, maintaining a better balance between survival and throughput across the entire 0.5~F to 10.0~F range. Beyond its hardware agnosticism, the \ac{ap} approach is particularly notable for its minimal computational footprint. While the \ac{rl} agents provide excellent adaptability, their deployment requires a more significant \ac{mcu} resource footprint due to the complexity of neural network inference. In contrast, the \ac{ap} approach operates as a mathematically direct, lightweight solver. Although it lacks the fine-grained tunability of the \ac{rl} framework, it provides a highly efficient alternative that achieves hardware-agnosticism with minimal computational and memory requirements.

\subsection{ITI vs. Off-Time Optimization}
The flexibility of the \ac{rl}-based approach allows for tailoring the system's behavior via different reward formulations. Comparing the \gls{iti}-optimized agent with the Off-Time-optimized agent reveals two distinct operational strategies.
The \gls{iti}-optimized agent is inherently designed to trade more total off-time in exchange for smaller task spacing and higher overall daily executions by taking more aggressively driven execution decisions. On the other hand, the Off-Time-optimized agent more strongly prioritizes survival. It minimizes blackout durations and entirely avoids power failures for all capacitor sizes starting from 2~F, but this extreme safety comes at the cost of executing fewer tasks and allowing a slightly higher average \gls{iti}. This comparison validates that the \gls{rl} framework can be effectively tuned according to the specific demands of the \ac{iot} application. In this work, we have explored agent optimizations for \gls{iti} and survival, but the \ac{rl}-based approach is definitely not limited to these configurations.

\section{Conclusion and Future Work}
\label{sec:conclusion_and_future_work}

This paper investigated the performance trade-offs of various hardware-agnostic, dynamic task execution strategies for batteryless, energy-harvesting \ac{iot} devices. By treating the application as a "black box," we evaluated a model-free \ac{rl} agent, an \acrlong{ap} method, an \acrlong{aimd}-based approach (AsTAR), and static baselines against a highly dynamic solar-harvesting environment.

A key takeaway from our extensive evaluation is that no single decision logic is universally superior; rather, each approach offers distinct benefits tailored to specific operational requirements. Our findings demonstrate that assessing the system's inherent hardware constraints is a critical first step in addressing energy volatility. When form-factor and economic constraints allow for a large energy buffer (e.g., > 5.5~F in our evaluated scenarios), simple and computationally inexpensive static thresholding policies are highly effective. In such less constrained systems, the advanced dynamic approaches evaluated in this work do not offer significant enough benefits to justify their inherent computational and memory overheads, serving as an important boundary regarding the practical utility of dynamic thresholding.

However, when strict hardware limitations force the system to operate with small capacitors, dynamic execution strategies become essential for maintaining continuous operation. Under these severe constraints, our proposed \ac{ap} approach provides a lightweight, real-time adaptation mechanism that closely matches optimal oracle throughput without requiring manual calibration. Alternatively, if the application requires a highly tunable balance between task execution and system survival, the \gls{rl}-based approach offers a flexible, hardware-agnostic policy. Finally, for applications that specifically require consistent execution pacing to bridge long harvesting gaps (such as night-time communication), AsTAR remains the most effective strategy.

Future work will focus on deploying these adaptive strategies onto physical hardware to validate the simulated energy overheads of the decision logic under real-world \gls{mcu} constraints. Additionally, exploring hybrid policies that seamlessly switch between static thresholding during peak harvesting hours and dynamic evaluation during energy-scarce periods presents a promising avenue for further optimizing batteryless, energy-neutral \ac{iot} systems.

\bibliographystyle{IEEEtran}

\bibliography{bibliography}

\begin{IEEEbiography}
[{\includegraphics[width=1in,height=1.25in,clip,keepaspectratio]{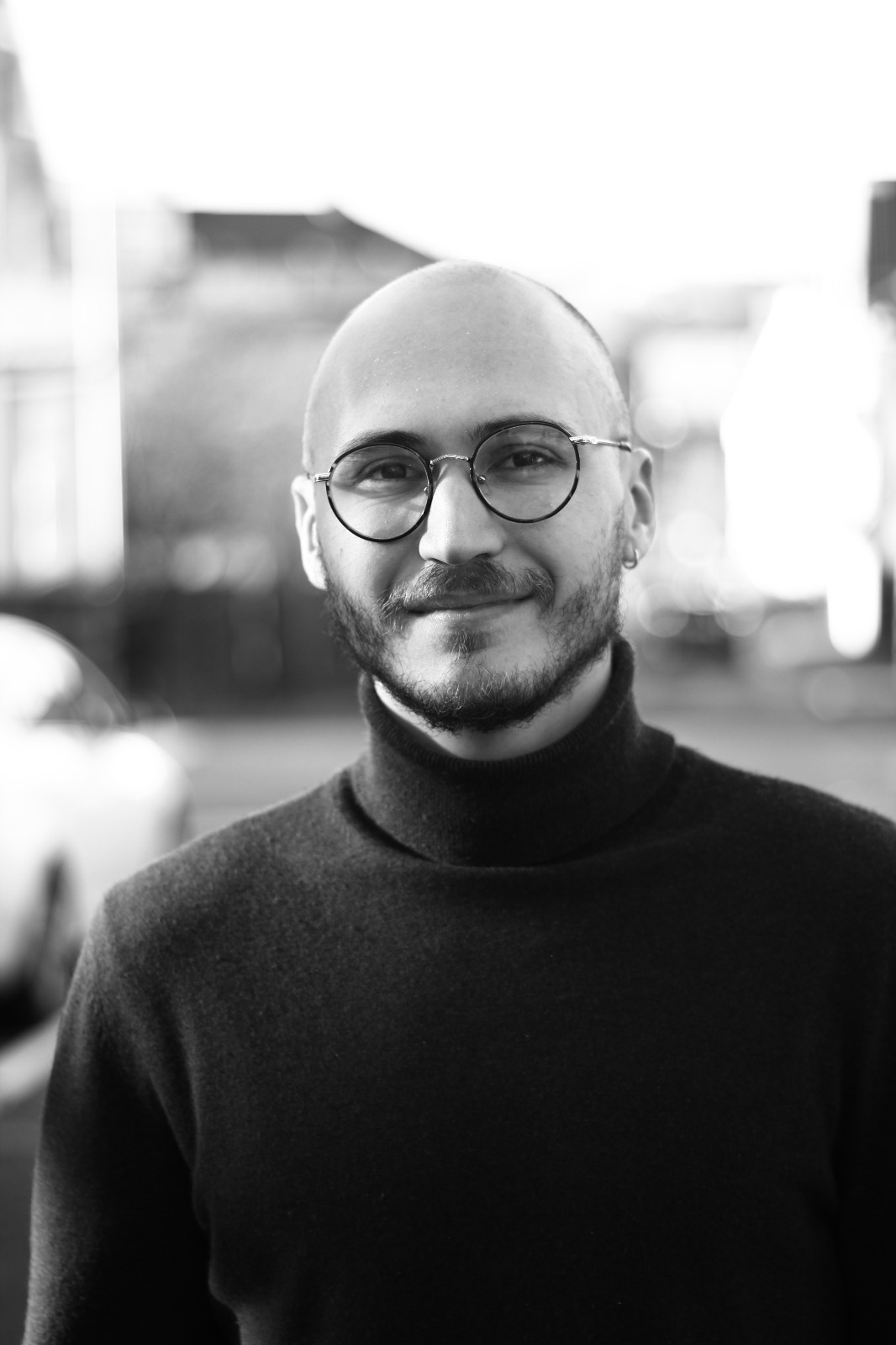}}]{Samer Nasser} Received his B.Sc. and M.Sc degrees in Electronics and ICT Engineering Technology from the University of Antwerp, Belgium, in 2020 and 2021, respectively. After working as an early-stage researcher in the field of environmental monitoring technology at DMR in Aalborg, Denmark, he is currently pursuing a Ph.D. in Applied Engineering at the University of Antwerp within the IDLab research group (University of Antwerp and IMEC). His work focuses on ambiently powered intelligent system design for sustainable IoT applications.
\end{IEEEbiography}

\begin{IEEEbiography}
[{\includegraphics[width=1in,height=1.25in,clip,keepaspectratio]{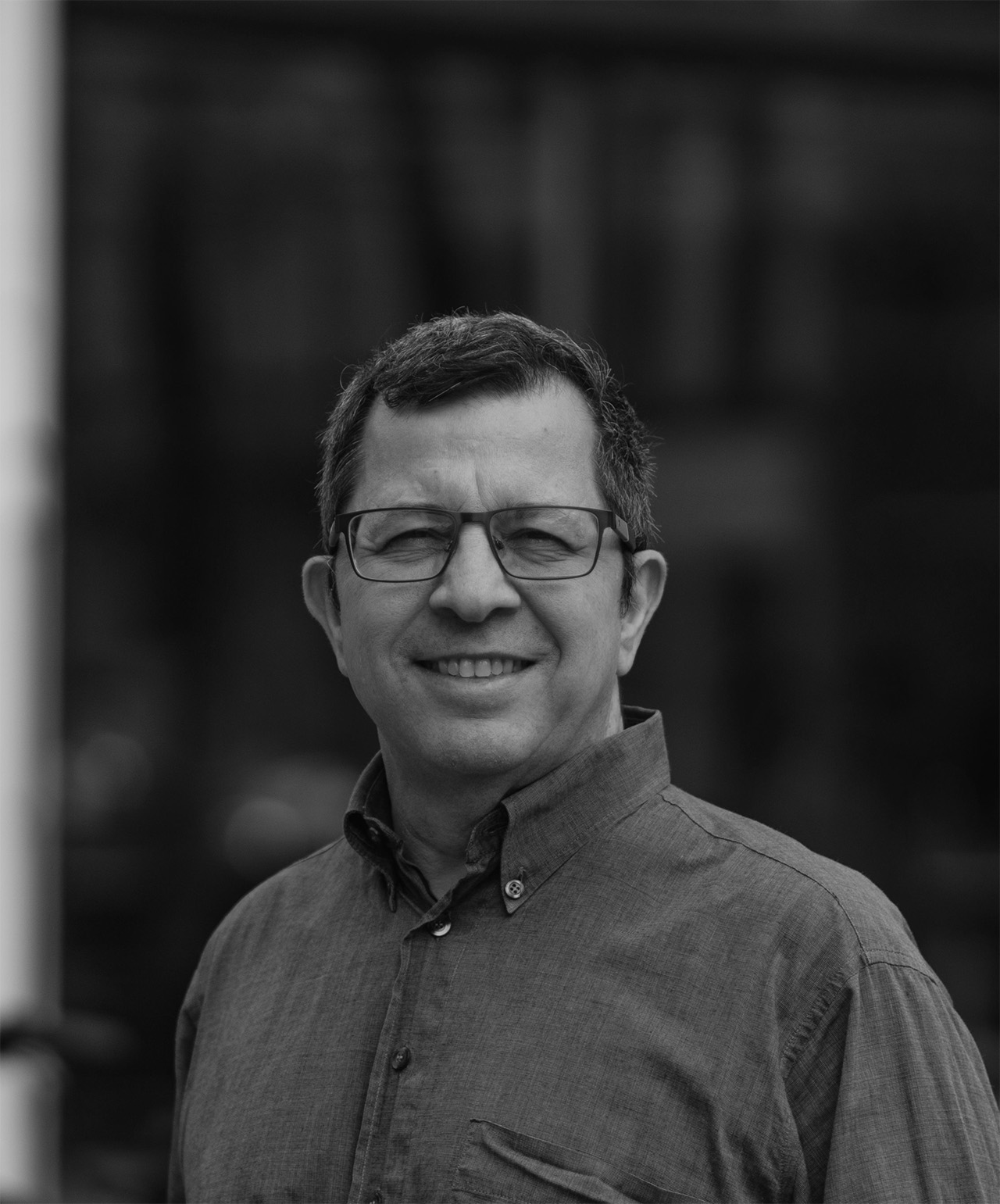}}]{Henrique Duarte Moura} is a senior researcher at the University of Antwerp and imec, Belgium.
He is a member of the Perceptive Radio Systems team in the IDLab research group.
He received his B.Sc degree in electrical and electronics engineering from the Universidade Federal de Minas Gerais (UFMG), Brazil, in 1991 and in system information from Universidade Estácio de Sá, Brazil, in 2012.
He received his M.Sc. and Ph.D. degrees in Computer Science from UFMG, Brazil, respectively, in 2015 and 2019.
\end{IEEEbiography}

\begin{IEEEbiography}
[{\includegraphics[width=1in,height=1.25in,clip,keepaspectratio]{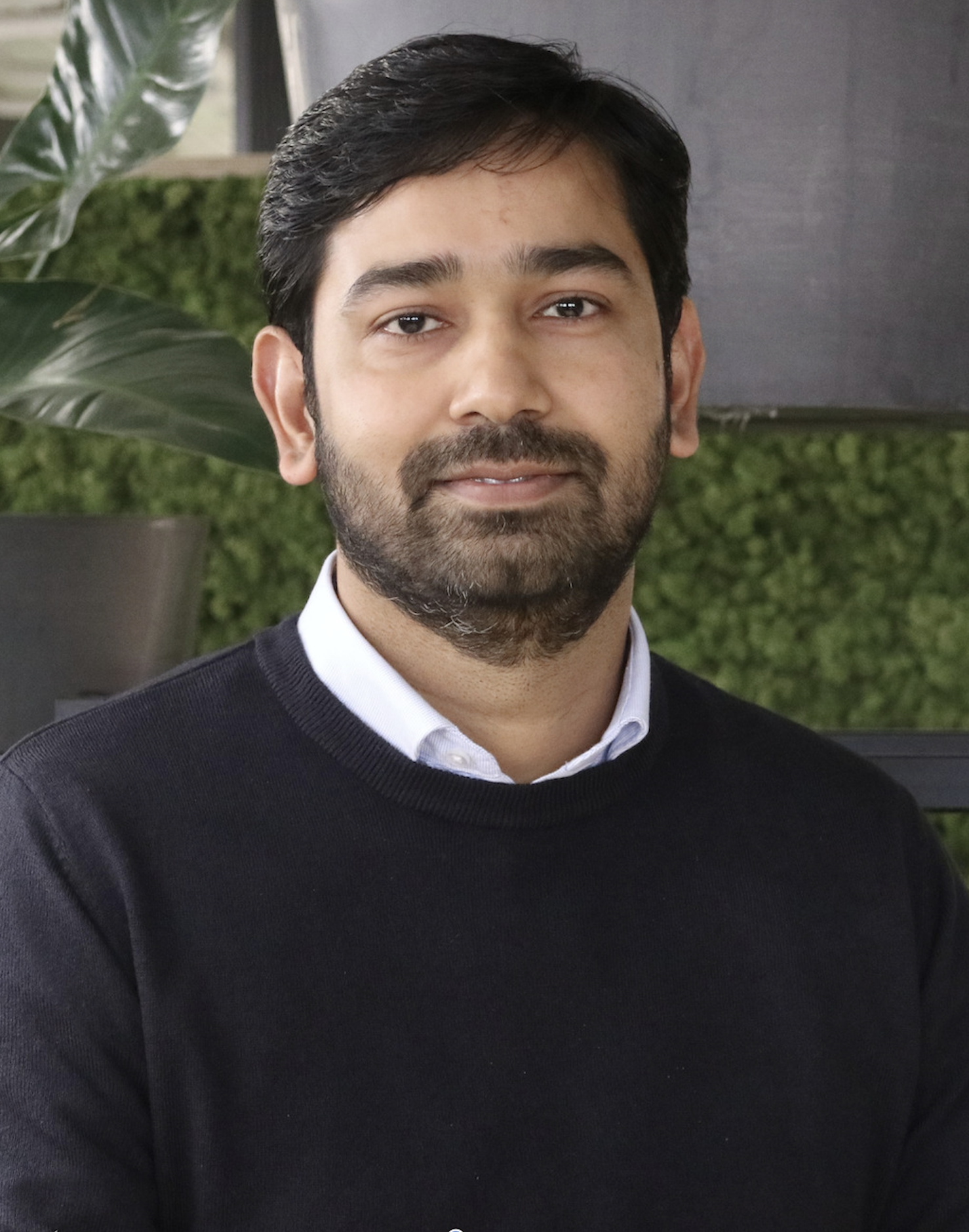}}]{Ritesh Kumar Singh} is a Principal Research Fellow at the University of Antwerp and a senior researcher at IMEC, Belgium. He is a member of the IDLab research group, where he leads the low-power portfolio. He received his M.Tech in Information Technology from IIIT, Allahabad, India in 2012 and subsequently worked in LG \& TCS research labs. He obtained his Ph.D. in Applied Engineering at the University of Antwerp in 2022. His current research interests include energy-aware computing, 6G, ML for low-power devices, and sustainable \gls{iot}.
\end{IEEEbiography}

\begin{IEEEbiography}
[{\includegraphics[width=1in,height=1.25in,clip,keepaspectratio]{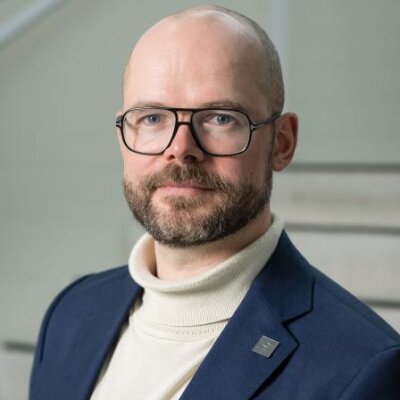}}]{Maarten Weyn} is a full professor and Vice-Rector of Research and Impact at the University of Antwerp. He teaches wireless communication system. His research at imec-IDLab focuses on ultra-low power sensor communication, embedded systems, sub-1 GHz communication, sensor processing, and localization. Maarten co-founded spin-offs Aloxy, CrowdScan, IoSa, and AtSharp, and contributed to 1OK and Viloc.
\end{IEEEbiography}

\begin{IEEEbiography}
[{\includegraphics[width=1in,height=1.25in,clip,keepaspectratio]{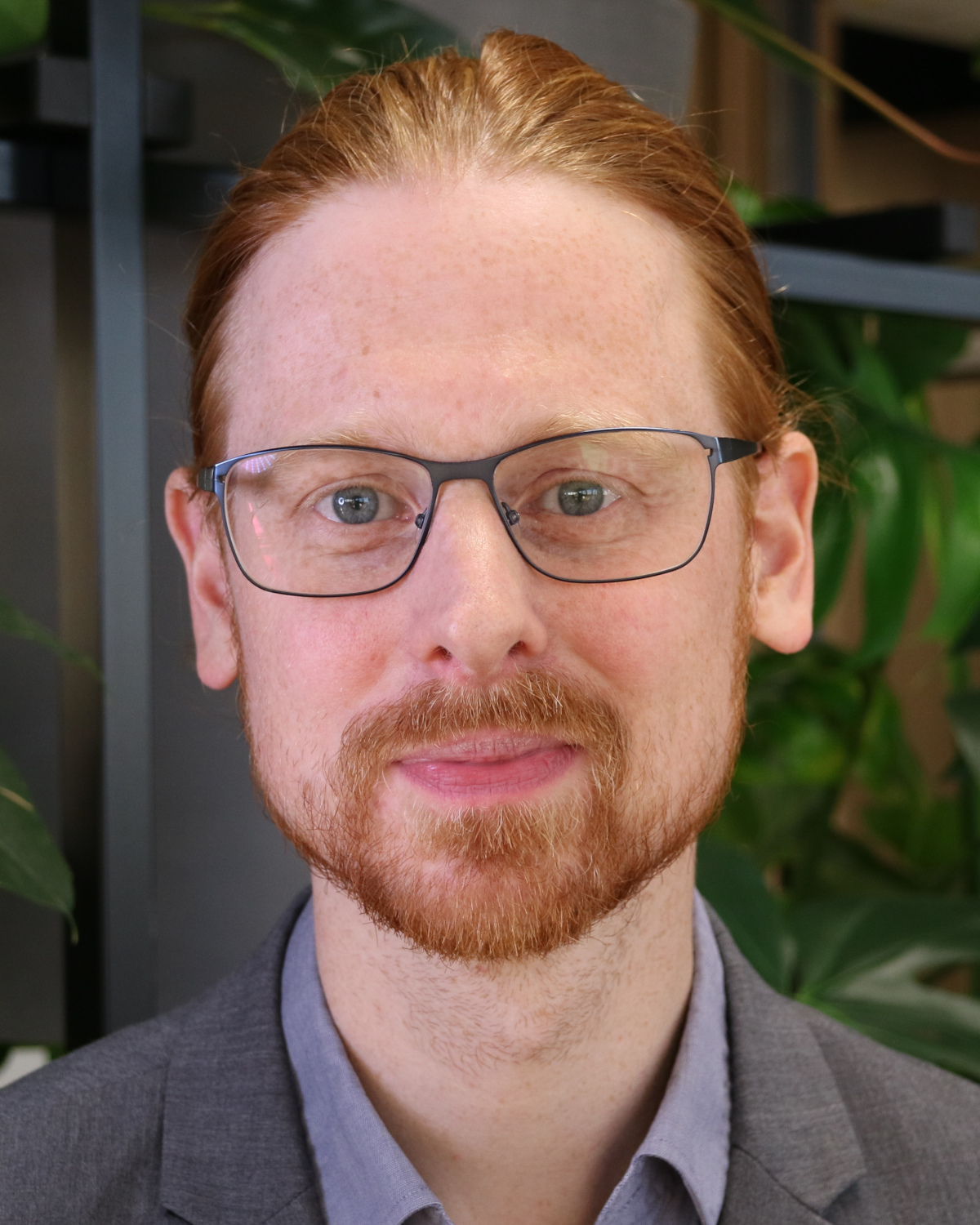}}]{Jeroen Famaey} is an associate professor at the University of Antwerp, Belgium, and a senior researcher at IMEC, Belgium. His current research interests include low-power distributed machine learning and wireless communications for Ambient IoT devices, as well as data-driven integrated sensing and communications. He has co-authored over 200 papers, published in international peer-reviewed journals and conference proceedings.
\end{IEEEbiography}

\end{document}